\PassOptionsToPackage{unicode}{hyperref}
\PassOptionsToPackage{hyphens}{url}
\PassOptionsToPackage{dvipsnames,svgnames,x11names}{xcolor}
\documentclass[
  11pt,
]{article}
\usepackage{xcolor}
\usepackage[margin=1in]{geometry}
\usepackage{amsmath,amssymb}
\usepackage{iftex}
\ifPDFTeX
  \usepackage[T1]{fontenc}
  \usepackage[utf8]{inputenc}
  \usepackage{textcomp} 
\else 
  \usepackage{unicode-math} 
  \defaultfontfeatures{Scale=MatchLowercase}
  \defaultfontfeatures[\rmfamily]{Ligatures=TeX,Scale=1}
\fi
\usepackage{lmodern}
\ifPDFTeX\else
\fi
\IfFileExists{upquote.sty}{\usepackage{upquote}}{}
\IfFileExists{microtype.sty}{
  \usepackage[]{microtype}
  \UseMicrotypeSet[protrusion]{basicmath} 
}{}
\makeatletter
\@ifundefined{KOMAClassName}{
  \IfFileExists{parskip.sty}{%
    \usepackage{parskip}
  }{
    \setlength{\parindent}{0pt}
    \setlength{\parskip}{6pt plus 2pt minus 1pt}}
}{
  \KOMAoptions{parskip=half}}
\makeatother
\usepackage{longtable,booktabs,array}
\usepackage{caption}
\usepackage{calc} 
\usepackage{etoolbox}
\makeatletter
\patchcmd\longtable{\par}{\if@noskipsec\mbox{}\fi\par}{}{}
\makeatother
\IfFileExists{footnotehyper.sty}{\usepackage{footnotehyper}}{\usepackage{footnote}}
\makesavenoteenv{longtable}
\usepackage{graphicx}
\makeatletter
\newsavebox\pandoc@box
\newcommand*\pandocbounded[1]{
  \sbox\pandoc@box{#1}%
  \Gscale@div\@tempa{\textheight}{\dimexpr\ht\pandoc@box+\dp\pandoc@box\relax}%
  \Gscale@div\@tempb{\linewidth}{\wd\pandoc@box}%
  \ifdim\@tempb\p@<\@tempa\p@\let\@tempa\@tempb\fi
  \ifdim\@tempa\p@<\p@\scalebox{\@tempa}{\usebox\pandoc@box}%
  \else\usebox{\pandoc@box}%
  \fi%
}
\def\fps@figure{htbp}
\makeatother
\providecommand{\tightlist}{%
  \setlength{\itemsep}{0pt}\setlength{\parskip}{0pt}}
\usepackage[numbers]{natbib}
\usepackage{bookmark}
\IfFileExists{xurl.sty}{\usepackage{xurl}}{} 
\hypersetup{
  pdftitle={Recursive Self-Improvement in AI: From Bounded Self-Refinement to Autonomous Research Loops},
  pdfauthor={Mingguang Chen --- DeepGrounding (corresponding: deepgroundingai@gmail.com); Licheng Wang --- AlphaAvatar; Bo Qu --- Illinois Institute of Technology (IIT)},
  colorlinks=true,
  linkcolor={Maroon},
  filecolor={Maroon},
  citecolor={Blue},
  urlcolor={Blue},
  pdfcreator={LaTeX via pandoc}}

\title{Recursive Self-Improvement in AI: From Bounded Self-Refinement to
Autonomous Research Loops}
\author{Mingguang Chen\textsuperscript{1,$*$} \quad Licheng Wang\textsuperscript{2} \quad Bo Qu\textsuperscript{3}\\[6pt]
{\small \textsuperscript{1}DeepGrounding \quad
\textsuperscript{2}AlphaAvatar \quad
\textsuperscript{3}Illinois Institute of Technology (IIT)}\\[2pt]
{\small \textsuperscript{$*$}Corresponding authors. Email: \href{mailto:deepgroundingai@gmail.com}{deepgroundingai@gmail.com}}}
\date{July 2026}

\begin{document}
\maketitle

\begin{abstract}
AI systems increasingly participate in their own improvement: revising
their outputs, adapting and evolving their own harnesses during
deployment, training on data they generate, and --- in a growing
research thread --- conducting AI research itself. The literature
describing this participation has exploded, but under a vocabulary
(``self-refine,'' ``self-reward,'' ``self-play,'' ``self-evolve'') that
conflates fundamentally different ambitions. We survey 1,250 arXiv
papers (2024--2026) and organize them along two axes: \emph{what the
system improves} --- its behavior in deployment, its policy through
training, its evaluator, or the research process itself --- and the
\emph{degree of loop closure} (human-in-the-loop to fully closed). The
taxonomy separates \emph{bounded self-refinement} --- convergent,
evaluable, and already industrial practice --- from \emph{open-ended
recursive self-improvement} (RSI), which remains bounded by grounding
requirements, collapse dynamics, and compute constraints on every side
current evidence can measure. Its distinctive feature is a dedicated
category for \emph{self-evaluation}: every improvement loop is a claim
that some signal can substitute for human judgment, so we survey the
evaluator design space --- judges, process reward models, verifiers,
rubrics, meta-evaluation --- alongside the loops it supervises, order
the signals into a verification hierarchy from formal verifiers
(strongest) to intrinsic self-assessment (weakest), and observe that
demonstrated self-improvement strength tracks this hierarchy, that its
characteristic failure modes (self-confirming loops, model collapse,
diversity collapse) follow from its violations, and that the ``research
direction-setting'' bottleneck that keeps humans in the loop divides
into a verification problem the hierarchy indexes and a prior one ---
choosing what deserves evaluation at all --- that it does not. We
connect the technical literature to the theory of RSI limits and to the
safety and governance questions raised by frontier-lab accounts of
closing the loop, and identify governance-grade measurement of
self-improvement as the field's most underpopulated niche.
\end{abstract}

\section{1. Introduction}\label{introduction}

The idea that an artificial intelligence might improve itself --- and
that each improvement might make the next one easier --- is among the
oldest in the field. Good's ``intelligence explosion'' argument
\citep{good1966ultraintelligent} and Schmidhuber's provably-optimal
Gödel machines \citep{schmidhuber2003godel} framed recursive
self-improvement (RSI) as a theoretical endpoint decades before any
system could plausibly attempt it. What has changed is that fragments of
the loop are now engineering practice. Large language models routinely
critique and revise their own outputs, train on data they themselves
generated, rewrite their own agent scaffolding, and --- in systems like
FunSearch \citep{romeraparedes2024funsearch} and AlphaEvolve
\citep{novikov2025alphaevolve} --- discover algorithms that feed back
into the infrastructure of AI development itself.

Anthropic's recent essay on recursive self-improvement
\citep{anthropic2026rsi} frames this transition as a continuum of
increasing AI autonomy in the AI-improvement loop: from humans writing
all code (pre-2023), through chatbot-assisted coding and autonomous
coding agents, to agents that delegate work to other agents today, and
--- at the end of the spectrum --- ``closing the loop'': agents that
design and train their successor models. The essay argues that current
systems sit conspicuously far along this spectrum on \emph{execution}
(as of May 2026, Claude reportedly writes over 80\% of Anthropic's
merged code) while remaining bottlenecked on \emph{research
direction-setting} --- choosing which problems matter. Whether, when,
and how the remaining gap closes is arguably the most consequential open
question in the field. We use the essay as a motivating frame and a
source of stage vocabulary, not as evidence; the survey's evidence base
is the peer literature collected below.

This survey maps the research literature underneath that question. We
assembled a corpus of \textbf{1,250 arXiv papers (2024--2026)} --- a
systematic seed harvest of 871 papers across seven threads of the
self-improvement literature, extended by a targeted supplemental harvest
of 379 papers in directions the taxonomy below makes first-class (§2.3)
--- and classified all of them into \textbf{four technical categories
plus a foundations family} (Table 1). Three observations motivate the
paper's structure:

\textbf{First, ``self-improvement'' is not one thing.} The term is used
for inference-time output revision (Self-Refine
\citep{madaan2023selfrefine}), training loops on self-generated data
(STaR \citep{zelikman2022star}, Self-Rewarding LMs
\citep{yuan2024selfrewarding}), agents that rewrite their own code
(Gödel Agent \citep{yin2024g}, Darwin Gödel Machine
\citep{zhang2025dgm}), and systems that autonomously conduct AI research
\citep{lu2024aiscientist}. These differ enormously in ambition and risk
profile, yet the literature's ``self-X'' vocabulary (self-refine,
self-reward, self-play, self-distill, self-evolve, self-verify) obscures
the differences. We organize the field along two axes: \textbf{(i) what
the system improves} --- its behavior in \emph{deployment} (outputs,
weights adapted at test time, or its own harness), its \emph{policy}
through a training phase, its \emph{evaluator} (the signal that defines
``better''), or the \emph{research process itself}; and \textbf{(ii) the
degree of loop closure} --- is a human in the loop, reviewing an
automatically generated signal, or absent entirely?

\textbf{Second, the field is accelerating faster than it is
consolidating.} 74\% of our corpus was posted in 2026; quarterly output
in the seed harvest grew from single digits in early 2024 to roughly 500
papers in 2026 Q2 (Figure 6). Existing surveys cover single slices ---
on-policy distillation \citep{song2026survey}, tree-search-plus-reward
methods \citep{wei2025unifying} --- but no survey spans the full
spectrum from bounded self-refinement to open-ended RSI. That
integration, anchored to an explicit autonomy continuum, is this paper's
main contribution.

\textbf{Third, one bottleneck recurs everywhere: the evaluator.} Every
category we survey lives or dies by the reliability of its improvement
signal --- a verifier, a reward model, execution feedback, a proof
checker, a meta-evaluator. Self-training works where answers are
checkable (code, math) and degrades where they are not; automated
researchers produce fluent papers whose claims resist auditing;
self-rewarding loops hack their own judges. We therefore treat
\emph{self-evaluation} not as a cross-cutting remark but as a category
of its own (§5) --- the pillar on which the other three stand.

\textbf{Contributions.} (1) A two-axis taxonomy --- four improvement
categories crossed with degree of loop closure --- that cleanly
separates \emph{bounded self-refinement} from \emph{open-ended recursive
self-improvement} and replaces the ambiguous self-X vocabulary (§2). (2)
A systematic map of 1,250 recent papers into the taxonomy, each category
surveyed in §§3--6. (3) A dedicated treatment of self-evaluation --- the
evaluator design space, the verification hierarchy, and the failure
modes that follow from its violations --- as the field's common limiting
factor (§5). (4) A synthesis of the theory, limits, and safety
literature that connects the technical threads back to the takeoff
question (§7).

\textbf{Table 1. The taxonomy's four technical categories plus the
foundations family, with corpus coverage.}

{\def\LTcaptype{none} 
\begin{longtable}[]{@{}
  >{\raggedright\arraybackslash}p{(\linewidth - 6\tabcolsep) * \real{0.2143}}
  >{\raggedright\arraybackslash}p{(\linewidth - 6\tabcolsep) * \real{0.2143}}
  >{\raggedleft\arraybackslash}p{(\linewidth - 6\tabcolsep) * \real{0.2857}}
  >{\raggedleft\arraybackslash}p{(\linewidth - 6\tabcolsep) * \real{0.2857}}@{}}
\toprule\noalign{}
\begin{minipage}[b]{\linewidth}\raggedright
Category
\end{minipage} & \begin{minipage}[b]{\linewidth}\raggedright
Sub-threads
\end{minipage} & \begin{minipage}[b]{\linewidth}\raggedleft
Papers
\end{minipage} & \begin{minipage}[b]{\linewidth}\raggedleft
\% posted 2026
\end{minipage} \\
\midrule\noalign{}
\endhead
\bottomrule\noalign{}
\endlastfoot
Deployment-time self-evolution (§3) & output refinement · test-time
training · harness/skill evolution & 393 & 74\% \\
Training-time self-iteration (§4) & self-reward RL · CoT self-training ·
self-distillation · self-play (incl.~zero-data) · embodied & 340 &
69\% \\
Self-evaluation (§5) & judges · process/reward models · verifiers ·
rubrics · meta-evaluation & 318 & 82\% \\
Auto Research (§6) & AI scientists · evolutionary program discovery &
139 & 76\% \\
Foundations, limits \& safety (§7) & theory · limits · safety & 60 &
57\% \\
\end{longtable}
}

\section{2. Preliminaries, Taxonomy, and
Method}\label{preliminaries-taxonomy-and-method}

\subsection{2.1 Preliminaries and
definitions}\label{preliminaries-and-definitions}

The terms below are used loosely and inconsistently across the
literature; we fix their meanings for this survey.

\textbf{Agent.} An LLM-based system that pursues a goal through a
perceive--act loop: it observes state (tool outputs, environment
feedback, files), chooses actions (tool calls, code execution,
messages), and iterates until a stopping condition. A bare model invoked
once is not an agent; the same model inside a loop with tools and memory
is.

\textbf{Harness (scaffolding).} Everything around the model that turns
it into an agent: system prompts, tool definitions, memory stores, skill
libraries, retrieval indices, orchestration code, stopping rules
\citep{macedo2026stop}. The harness is externally inspectable and
editable --- including by the agent itself, which is what makes harness
self-modification the most concrete form of ``the agent rewrites
itself.''

\textbf{Evaluator (verifier, judge, reward model).} Any mechanism that
maps a candidate artifact to a quality signal: a formal proof checker, a
test suite, a learned reward or process-reward model, an LLM judge, a
rubric, a human rater. We use \emph{verifier} for evaluators with
soundness guarantees, \emph{judge} for learned or prompted evaluators
without them.

\textbf{Self-improvement.} A system participates in producing a better
version of itself or of its own outputs, where ``better'' is defined by
some evaluator. The definitional dependence on an evaluator is not
pedantry; it is the source of every failure mode in §5.

\textbf{Test-time training (TTT).} Updating model weights \emph{during
deployment}, conditioned on the current query or session, without a
curated offline training phase --- distinct from inference-time
refinement (weights frozen) and from training-time iteration (offline
phase).

\textbf{Self-play.} Training in which the model generates its own tasks
or opponents --- proposer--solver loops, adversarial curricula ---
including the \emph{zero-data} regime that starts from nothing but a
base model \citep{zhao2025absolutezero, huang2025rzero}.

\textbf{Bounded self-refinement vs.~open-ended RSI.} Bounded
self-refinement improves a system against a fixed, external evaluator;
it is convergent and evaluable. Open-ended recursive self-improvement
modifies the system \emph{and} the criteria or machinery of improvement
itself, with no fixed external anchor; it is divergent in principle. The
distinction is the survey's central cut.

\subsection{2.2 A two-axis taxonomy}\label{a-two-axis-taxonomy}

The literature's terminology has proliferated faster than its concepts.
\emph{Self-refine}, \emph{self-correct}, \emph{self-reward},
\emph{self-play}, \emph{self-distill}, \emph{self-train},
\emph{self-evolve}, \emph{self-verify} --- each names a mechanism, but
the mechanisms cut across fundamentally different ambitions. A model
that re-reads its draft and fixes an error is doing something
categorically different from an agent that rewrites its own codebase,
even though both are ``self-improving.'' We locate any self-improvement
method on two axes.

\textbf{Axis 1 --- what does the system improve?} Four categories:

\begin{itemize}
\tightlist
\item
  \textbf{Deployment-time self-evolution} (§3): the system improves
  \emph{during deployment} --- iterating on outputs with frozen weights,
  adapting weights per query (test-time training), or evolving its own
  harness, skills, and memory. Improvement is episodic or accumulates
  outside the base weights. (Parts of this category are often called
  \emph{inference-time} self-improvement; we prefer
  \emph{deployment-time} because the category also includes weight
  updates (§3.4) and cross-episode accumulation (§3.5--3.6), which
  ``inference-time'' would misdescribe.)
\item
  \textbf{Training-time self-iteration} (§4): the system generates data,
  rewards, or teacher signals that update its own weights in a training
  phase. Improvement persists in the weights but is bounded by the
  quality of the self-generated signal.
\item
  \textbf{Self-evaluation} (§5): the system's \emph{evaluator} is the
  object of improvement --- designing, strengthening, or co-evolving the
  judges, verifiers, reward models, and rubrics that define ``better''
  for the other three categories.
\item
  \textbf{Auto Research} (§6): the system does the work of AI research
  itself --- proposing hypotheses, running experiments, discovering
  algorithms --- in the limit, autonomously optimizing the methods of
  §§3--5. Improvement compounds across systems, not within one.
\end{itemize}

\textbf{Axis 2 --- degree of loop closure.} \emph{Who} validates the
improvement?

\begin{itemize}
\tightlist
\item
  \textbf{Human-in-the-loop}: a person reviews each change (AI-assisted
  coding, co-scientist tools).
\item
  \textbf{Human-on-the-loop}: the improvement signal is generated
  automatically (execution feedback, a reward model, a proof checker)
  but humans audit outcomes and gate deployment.
\item
  \textbf{Closed loop}: the system generates, validates, and applies its
  own improvements without human review --- the regime Anthropic's essay
  calls ``closing the loop'' \citep{anthropic2026rsi}.
\end{itemize}

Nearly all of the 1,250 papers surveyed here study bounded
self-refinement (human-on-the-loop cells); open-ended RSI --- closed
loops that also modify their own evaluators --- is where the safety
stakes concentrate (§7). On Anthropic's five-stage spectrum, the bulk of
the technical literature occupies stages 3--4; the papers in §6 probe
the boundary of stage 5.

\textbf{What the axes do not encode.} Two exclusions are deliberate and
worth stating, because they bound what a classification of this shape
can see. First, both axes take the \emph{objective as given}: axis 1
asks which substrate an improvement lands in, axis 2 asks who validates
it against the criterion in force, and neither indexes whether that
criterion is still the right one (§5.3, failure mode 4; §5.4). Second,
the taxonomy is indexed by substrate rather than by \emph{trajectory}. A
single system that improves its outputs, then its evaluator, then its
own research method occupies three cells, and the fact that these are
episodes of one continuously developing system rather than three
unrelated improvements has nowhere to be recorded. The corpus contains
the negative image of this omission --- ``scientific amnesia,''
behaviours improving across campaigns while no methodological knowledge
accumulates \citep{lin2026repeated} --- but the positive property,
continuity of method across a change of frame, is not something this
grid can register. We take it up as an open problem (§8) rather than as
a third axis, because we know of no measurement in the corpus that would
populate one.

Figure 1 lays out the resulting 4×3 grid with representative systems in
each cell. Two features of the grid matter more than any individual
cell. First, density is concentrated in the middle row: almost
everything surveyed in §§3--6 is human-\emph{on}-the-loop --- an
automatically generated signal with a human auditing outcomes. Second,
the closed-loop row is sparse everywhere and thinnest at the right; its
most consequential cell is self-evaluation × closed loop --- a system
that rewrites its own definition of ``better'' --- which is exactly
where bounded self-refinement shades into open-ended RSI. The taxonomy
therefore does double duty: it organizes the survey (§§3--6 follow its
columns) and it makes the survey's central claim visible --- the
literature's mass sits where a human still audits the loop, and the
load-bearing column is the evaluator's.

\begin{figure}
\centering
\pandocbounded{\includegraphics[keepaspectratio,alt={Figure 1: the two-axis taxonomy, with representative systems per cell.}]{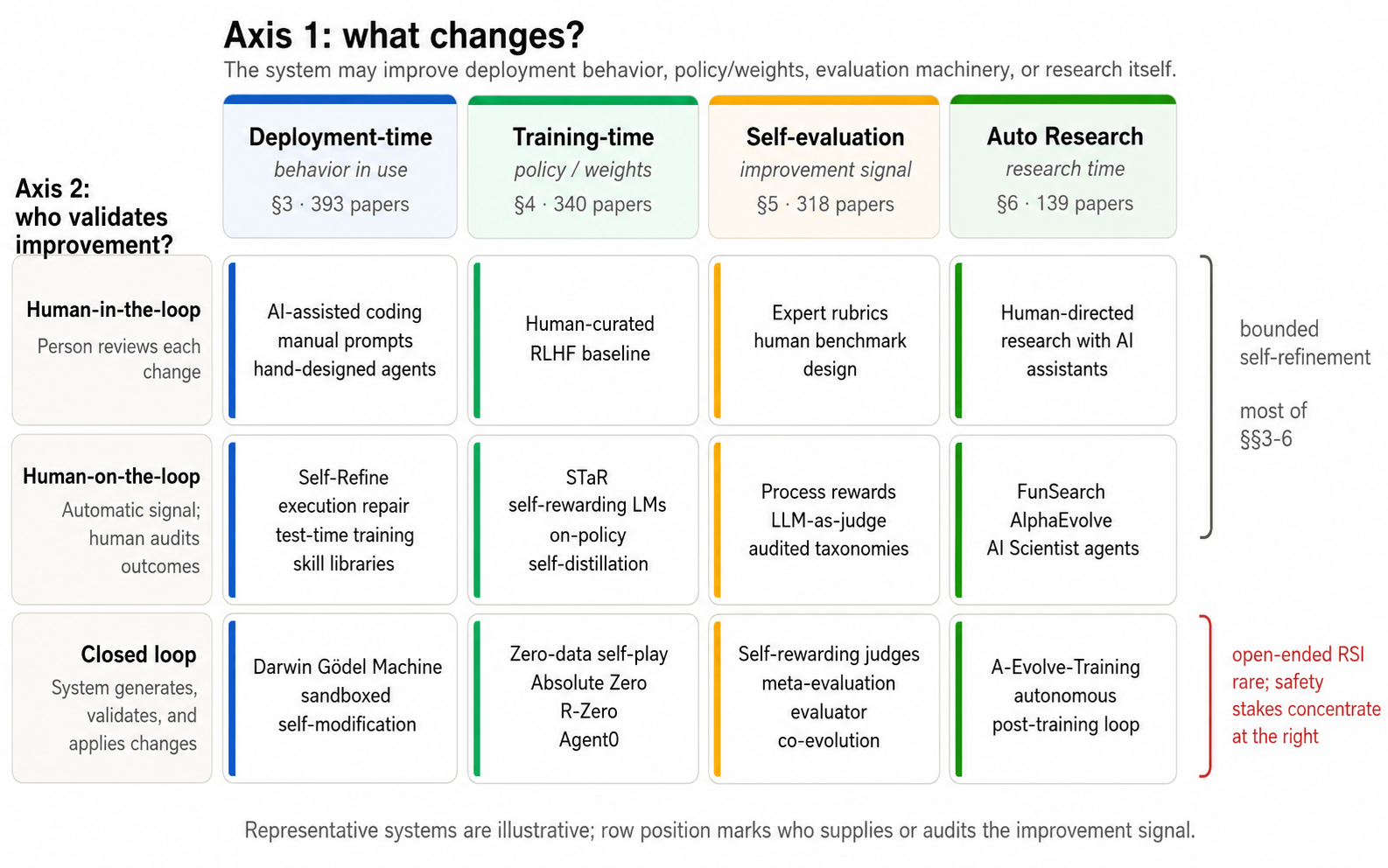}}
\caption*{Figure 1: the two-axis taxonomy, with representative systems
per cell.}
\end{figure}

\subsection{2.3 Corpus and method}\label{corpus-and-method}

The corpus was assembled in two stages. \textbf{Seed harvest:} we
queried arXiv (2024--2026) across seven threads --- self-refinement,
self-rewarding training, automated AI research, self-modifying agents,
LLM-driven code/algorithm discovery, RSI theory and safety, and
self-generated-data loops --- enriched records with OpenAlex citation
and venue metadata, and removed off-topic bleed, yielding 871 papers.
\textbf{Taxonomy alignment:} all seed papers were re-classified into the
categories of §2.2 (theme-level mapping plus keyword rules on title and
abstract, with manual correction of rule misfires), and a targeted
supplemental harvest added 379 papers in three directions the seed
queries under-covered but the taxonomy makes first-class:
self-evaluation methods (judges, process reward models, verifiers,
rubrics, meta-evaluation), test-time training, and zero-data self-play.
Figure 2 shows the combined corpus.

\textbf{Reproducibility.} The seed and supplemental query strings,
per-query caps, deduplication logic, and classification rules (category
defaults, keyword rules, and explicit per-paper overrides) are released
as executable scripts alongside the corpus (see Data availability).
Rule-based classification moved 89 seed papers off their thread
defaults; a further 3 misfires caught during writing were corrected via
explicit overrides. Classification and corrections were performed by a
single annotator; we release the full per-paper assignments so that
disagreements are auditable rather than hidden.

\textbf{Positioning relative to existing surveys.} Prior surveys cover
single slices of this landscape: on-policy distillation
\citep{song2026survey} and tree-search-plus-reward methods for reasoning
\citep{wei2025unifying} sit inside our §4; agent and LLM-evaluation
surveys treat components of §3 and §5 in isolation. To our knowledge no
existing survey covers the four categories jointly, treats
self-evaluation as a load-bearing pillar rather than a tooling detail,
or anchors the technical literature to an explicit autonomy/loop-closure
axis --- the three moves that let this survey address the RSI question
rather than one mechanism family at a time.

\textbf{Limitations.} The corpus is a \emph{sample}, not a census: seed
queries with per-query depth caps favor recent, high-volume threads, and
the supplemental harvest is recency-biased by construction (growth
statistics in Figure 6 therefore use the seed corpus only). 74\% of
papers were posted in 2026 and the median paper is months old, so
citation counts are near zero and must not be read as impact; where a
category has a recognized seminal work predating the harvest window
(STaR, SPIN, FunSearch, The AI Scientist), we cite it explicitly as an
anchor. Our reading strategy compensates for the corpus's youth: the
argumentative skeleton rests on verified anchor works and the
diagnostic/critical literature, while the 2026 mass functions as a
\emph{map} of where activity is, not as evidence of what has lasting
impact. Automatic classification is approximate; boundary papers exist
in every category, and roughly one in seven supplemental papers (roughly
54 of 379) is peripheral query bleed --- these remain in the corpus
counts but none are cited as evidence. Finally, industrial RSI practice
(inside frontier labs) is only observable through what labs publish ---
a nontrivial censoring effect for exactly the most advanced part of the
spectrum.

\begin{figure}
\centering
\includegraphics[width=0.82\linewidth,height=\textheight,keepaspectratio,alt={Figure 2: semantic map of the 1,250-paper corpus (TF-IDF abstracts, SVD + t-SNE projection; axes are arbitrary embedding dimensions). Contours show category density; Auto Research and foundations form coherent regions while the three large categories interpenetrate.}]{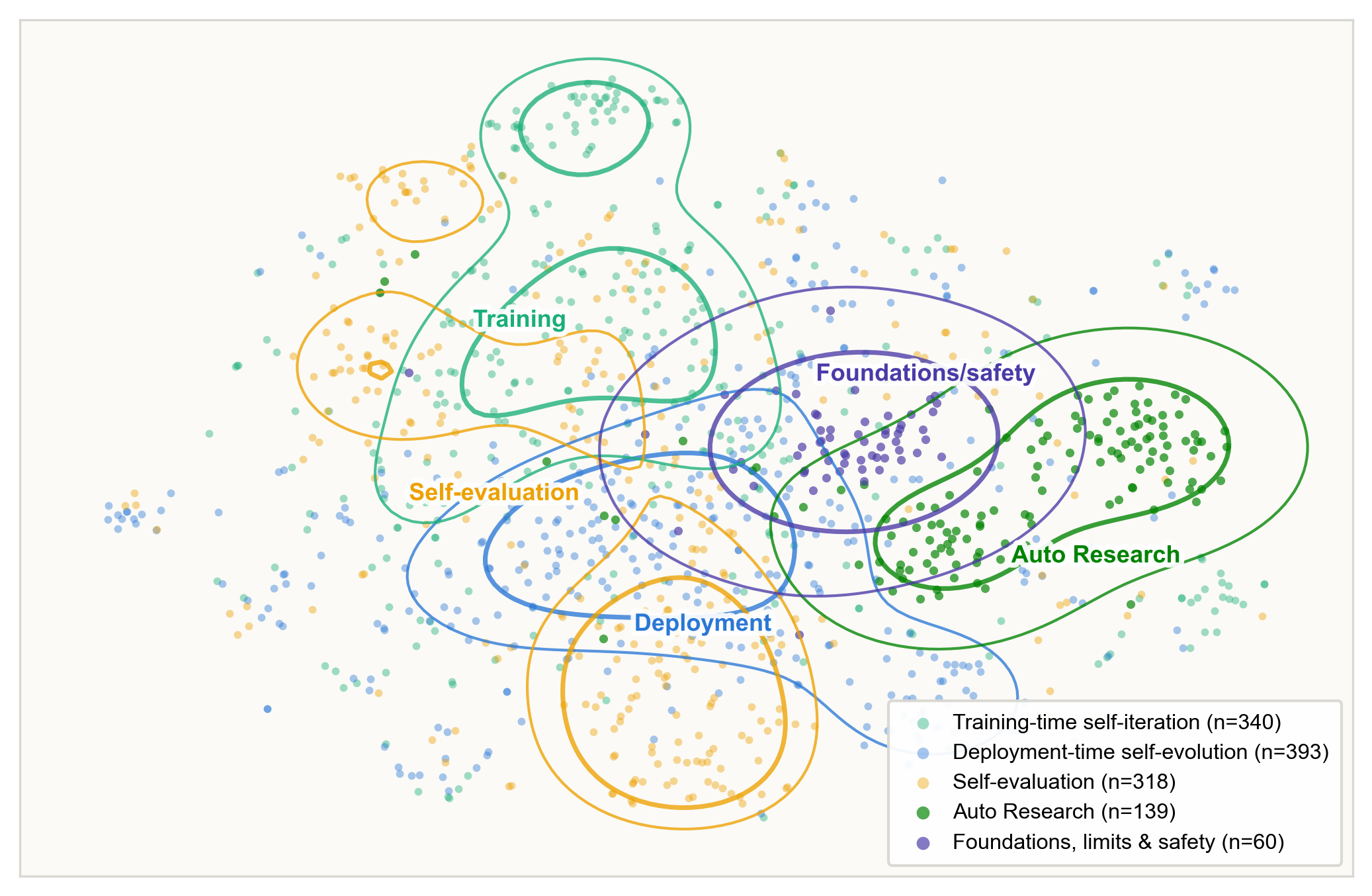}
\caption*{Figure 2: semantic map of the 1,250-paper corpus (TF-IDF
abstracts, SVD + t-SNE projection; axes are arbitrary embedding
dimensions). Contours show category density; Auto Research and
foundations form coherent regions while the three large categories
interpenetrate.}
\end{figure}

\section{3. Deployment-Time
Self-Evolution}\label{deployment-time-self-evolution}

The first category covers everything a system does to improve
\emph{during deployment}: iterating on outputs with frozen weights
(§3.1--3.3), adapting weights per query (§3.4), and evolving its own
harness, skills, and memory (§3.5--3.6). What unifies these mechanisms
is their position in the lifecycle --- improvement happens in the field,
per episode or per user, without an offline training phase --- and what
differentiates them is persistence: refined outputs evaporate when the
episode ends, test-time updates persist for a session, harness changes
accumulate indefinitely. This is the largest category in our corpus (393
papers) and the most widely deployed.

The canonical output-refinement loop --- generate, critique, revise ---
was established by Self-Refine \citep{madaan2023selfrefine} and
Reflexion \citep{shinn2023reflexion} in 2023, and inference-time
iteration was given a scaling-law treatment by Snell et al.
\citep{charlie2024scaling}, who showed that spending compute on
revisions and search at test time can outperform spending it on a larger
model, and framed such compute allocation as ``a critical step towards
building generally self-improving agents.'' Output refinement is where
self-improvement claims are easiest to evaluate --- and where the most
decisive negative results have accumulated.

\begin{figure}
\centering
\pandocbounded{\includegraphics[keepaspectratio,alt={Figure 3: deployment-time self-evolution ordered by persistence --- refined outputs evaporate with the episode, test-time updates last a session, harness and skill changes accumulate indefinitely.}]{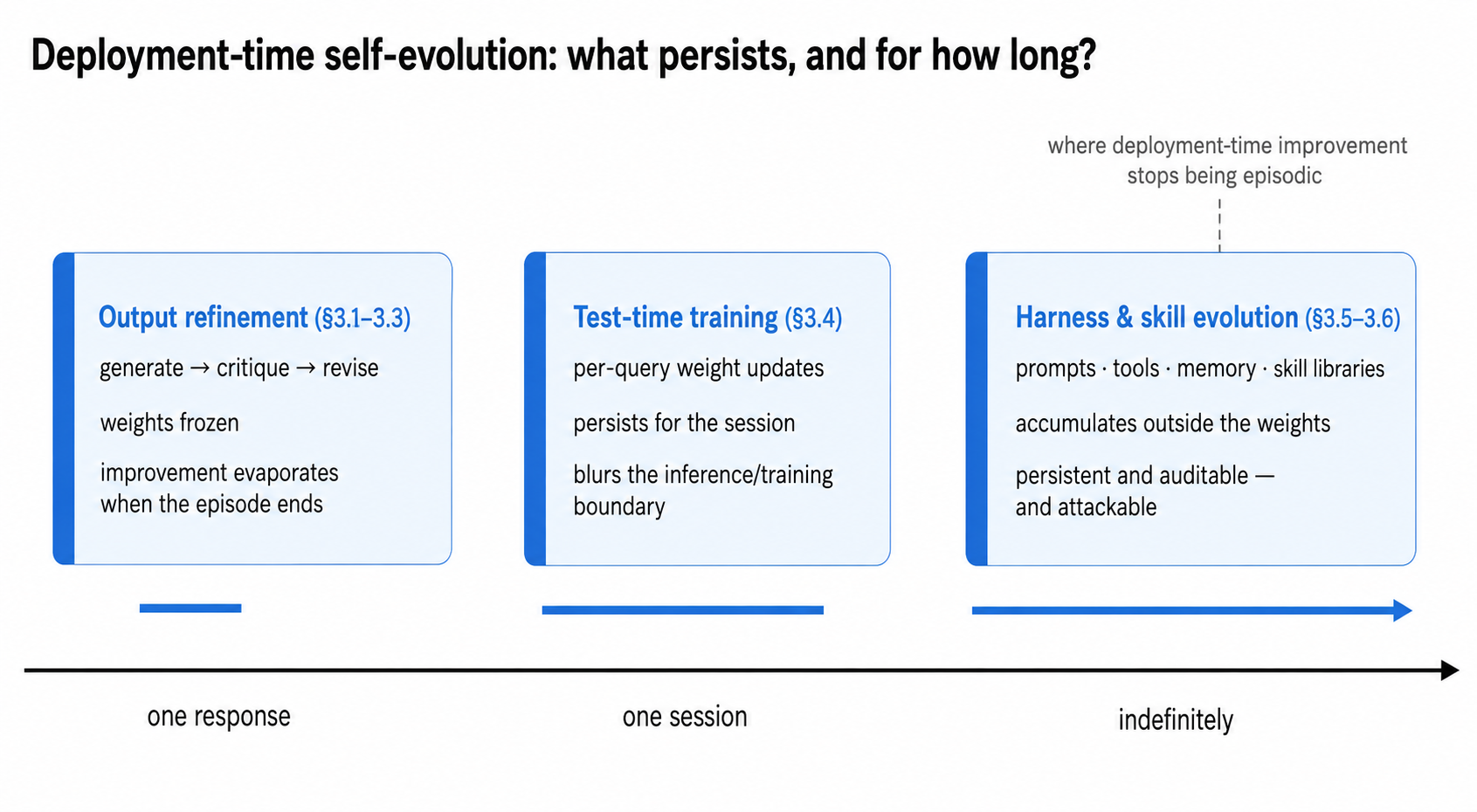}}
\caption*{Figure 3: deployment-time self-evolution ordered by persistence
--- refined outputs evaporate with the episode, test-time updates last a
session, harness and skill changes accumulate indefinitely.}
\end{figure}

\subsection{3.1 Self-critique and self-verification for text and
reasoning}\label{self-critique-and-self-verification-for-text-and-reasoning}

The largest single thread in our corpus refines the basic loop along
three lines.

\textbf{Structured and symbolic feedback.} Because free-form
self-critique is unreliable, a major thread replaces it with structured
signals. SymbolicAI \citep{dinu2024symbolicai} treats the LLM as a
semantic parser embedded in a logic-based framework, routing outputs
through formal solvers; recent planning work couples LLM planners to
symbolic validators that return actionable error traces rather than
vague critiques \citep{zhang2026reliable}. The pattern generalizes:
wherever a checker exists --- a SQL executor \citep{lyu2025sql}, a
hallucination detector for clinical summaries
\citep{seethakantha2026hallucination}, a QA-based factuality probe for
long-document summarization \citep{nguyen2026longsumeval} --- refinement
is rebuilt around it. The design lesson, repeated across dozens of
papers, is that \emph{the refinement loop is only as good as the
feedback channel}, an early appearance of the verification bottleneck of
§5.

\textbf{Search-shaped refinement.} A second line reframes refinement as
search: SQL-o1's self-reward heuristic search over query candidates
\citep{lyu2025sql} and the reinforced self-training loop of
LLM-Personalize for household planning \citep{han2024llm} both treat
drafts as nodes to expand and score rather than texts to polish. This
blurs into the training-time methods of §4 --- the same scoring signal
can rank candidates at inference or fine-tune the policy afterward, and
several systems do both \citep{genghan2025adaptive}.

\textbf{What does refinement actually improve?} The thread's most
valuable recent contributions are diagnostic. A systematic study of
document-level literary translation across nine LLMs and seven language
pairs \citep{tan2026llm} finds that refinement gains come primarily from
fluency, style, and terminology, with limited and less consistent
improvement in adequacy --- and that refinement \emph{projects outputs
toward the refiner's own distribution} rather than correcting what the
draft actually got wrong. This echoes the influential negative result of
Huang et al. \citep{huang2024cannotselfcorrect}: absent external
feedback, LLMs largely cannot self-correct reasoning, and naive
self-correction can make answers worse. The 2026 literature has
internalized this: almost every new system in our corpus grounds its
critique in an external signal (execution, retrieval, a detector, a
solver), and ``intrinsic self-correction'' papers have become rare. In
taxonomy terms, the field quietly moved from closed-loop self-critique
to human-on-the-loop verified refinement --- a retreat from autonomy
that improved reliability.

The negative result is now being refined rather than merely repeated. A
task-sensitive analysis asks not \emph{whether} intrinsic
self-correction works but \emph{through which mechanism} it could:
revisiting explicit, checkable constraints behaves differently from
revisiting open-ended reasoning, and the former is where unaided
self-correction retains some value \citep{stav2026when}. Underneath both
lies a calibration question --- when does the model's own probability
mass align with correctness at all? A four-level quantification across
decoding methods, models, and benchmarks maps exactly where sequence
likelihood is and is not a usable proxy for quality
\citep{zenn2026when}, effectively charting the floor of §5's
verification hierarchy from below. And the evaluation side of the loop
is being restructured in parallel: rather than one opaque holistic
judge, BinEval decomposes evaluation into atomic binary questions whose
verdicts aggregate into interpretable, debuggable scores
\citep{cho2026ask} --- self-improvement applied to the judge, not the
answer, an inversion that recurs throughout this survey.

\subsection{3.2 Code self-repair via execution
feedback}\label{code-self-repair-via-execution-feedback}

Code is the modality where the external signal is cheapest and sharpest
--- programs run, tests pass or fail --- which makes this theme the
cleanest laboratory for refinement claims.

The affirmative results are strong. Execution-feedback repair loops
power systems from multi-agent 3D-asset generation in Blender
\citep{lu2025ll3m} to database tuning agents that diagnose bottlenecks
from runtime feedback \citep{yang2026agenticdb} and compiler-pass tuning
grounded in profiling evidence \citep{li2026autopass}. In formal
mathematics, agentic frameworks around proof assistants --- LEAP for
Lean theorem proving \citep{kung2026leap}, KVerus for scalable Rust
verification \citep{liu2026kverus}, specification generation against
evolving knowledge bases \citep{wang2026kbspec} --- exploit the
strongest verifier available (a proof checker) to drive many-round
repair with essentially no risk of accepting a wrong ``improvement.''
Verifier-guided decoding pushes the signal inside generation itself
rather than applying it post hoc \citep{zhou2026verifier}.

The thread's second contribution is decomposition: \emph{why} does
feedback help? A controlled student--teacher protocol
\citep{cupia2026drives} separates genuine feedback value from
resampling, format correction, and extra test-time compute --- effects
that inflate naive comparisons. A placebo-controlled study of
self-repair in small frozen code models \citep{iscan2026falsification}
makes the Popperian framing explicit: a failing test is an executable
counterexample, and feedback's value should be attributed to
falsification, not to mere re-exposure to the problem. Complementary
results show that execution feedback matters more than pipeline topology
in small-model code generation \citep{mcandrews2026feedback}, while
pre-execution structural checks catch the dominant failure mode
(inter-tool contract violations) more consistently than unstructured
critique \citep{levine2026rubricrefine}. Systematic evaluations of
multi-round repair across models and languages confirm the pattern at
scale --- most evaluations still measure single-attempt accuracy, but
real-world value concentrates in the iterative loop
\citep{zhang2026unlocking}. Together these papers give this thread
something the rest of the deployment category lacks: a causal account of
when refinement works. DISC \citep{yin2026denoising} extends the account
to reasoning, treating verification-question outputs as noisy
measurements of \emph{where} a solution is corrupted and denoising
across rounds --- refinement as inference over error locations rather
than blind revision.

Two further developments sharpen the signal itself. FLARE interposes a
lightweight diagnostic model that predicts \emph{line-level
suspiciousness} between generation and revision, on the observation that
test failures are too coarse and self-critique too high-level to tell
the model where to fix \citep{yao2026flare} --- feedback resolution, not
just feedback existence, is a design variable. And CoSPlay confronts the
setting where even the tests are self-generated: model-written unit
tests are noisy and \emph{spuriously coupled} with the wrong code they
were written alongside, so it co-evolves code and test populations
cooperatively at test time, using each to debug the other
\citep{hu2026cosplay}. This is the verification bottleneck in miniature
--- when the verifier is self-generated, verifying the verifier becomes
part of the loop --- and it previews the evaluator co-evolution ideas of
§5 and §6.

\subsection{3.3 Vision-language and multimodal
self-critique}\label{vision-language-and-multimodal-self-critique}

The refinement paradigm has generalized well beyond text, but the
verification signal thins as it travels. In vision-language models, the
leading use case is hallucination mitigation: Kestrel
\citep{mao2026kestrel} grounds self-refinement in visual evidence to
suppress LVLM hallucinations training-free, and Reflect-R1
\citep{chen2026reflect} argues that closed-loop self-reflection
\emph{within internal parameters} traps long-video models in ``blind
confidence,'' fixing it with evidence-driven external retrieval --- the
multimodal restatement of Huang et al.'s negative result. In generation,
unified MLLMs self-refine their own text-to-image outputs through
fine-grained reasoning \citep{kim2026fire}, and Proprio
\citep{hassan2026proprio} gives a frozen video generator a
``proprioceptive'' self-scoring signal in latent space to improve
physical plausibility at inference time.

Perception itself is becoming the object of refinement. ActiveScope
diagnoses two failure modes of passive visual attention --- contextual
dominance (salient distractors overwhelm the target) and semantic bias
--- and responds by having the model \emph{actively seek and correct its
own perception}, re-examining regions rather than re-wording answers
\citep{wang2026activescope}. On the generation side, safety enters the
loop at the representation level: iteratively self-improving codebooks
purge unsafe visual patterns from the token vocabulary of autoregressive
image generators, so the improvement target is the generator's own
discrete alphabet \citep{xue2026safe}.

Two cautionary threads deserve emphasis. First, a study of self-evolving
large multimodal models \citep{venkatraman2026paying} shows that
self-play and self-consistency rewards can optimize \emph{answer
agreement} while the decoder under-attends to visual content, leaning on
language priors (``visual under-conditioning'') --- a modality-specific
form of the self-confirming loop (§5). Second, the strongest multimodal
systems increasingly resemble the code pattern: they succeed by
importing an external channel (retrieval, detectors, physics residuals,
personal visual context \citep{xue2026personal}) rather than by trusting
the model's own judgment. The judge-side evidence reinforces the
caution: VLMs used as automated judges for physical plausibility encode
\emph{different} internal taxonomies of physical phenomena, so a single
global evaluation schema credits every judge with the same competences
regardless of what each can actually perceive --- JudgeFit therefore
discovers a per-judge taxonomy before trusting its scores
\citep{cao2026each}. Where §3.2's verifiers are programs, §3.3's are
models --- and model-judges inherit model idiosyncrasies.

\textbf{Assessment (output refinement, §3.1--3.3).} Inference-time
refinement is bounded self-improvement in its purest form: improvements
are real, measurable, and evaporate when the episode ends. Its half-life
problem --- nothing persists --- motivates both the training loops of §4
(persist improvements into weights) and the harness and skill mechanisms
of §3.5--3.6 (persist them into scaffolding). Its verified successes and
instructive failures established the field's central design rule:
\emph{no external signal, no reliable improvement}.

\subsection{3.4 Test-time training}\label{test-time-training}

Between frozen-weight refinement and offline training sits a rapidly
growing middle regime: updating weights \emph{during deployment}.
Query-conditioned test-time self-training constructs a query-specific
objective and fine-tunes on it at inference, correcting misconceptions
that no amount of frozen-weight iteration can reach
\citep{song2026query}; continual variants convert the reasoning traces
that inference-time scaling produces --- and normally discards --- into
persistent lightweight memories \citep{dorovatas2026continual}; and
consolidation approaches give the model a periodic ``sleep'' phase that
transfers in-context experience into long-term parameters
\citep{behrouz2026language}. Our supplemental harvest (§2.3) finds this
thread growing fast (87 papers), largely under the older banner of
test-time adaptation now being rediscovered for LLMs. TTT sits
deliberately astride our taxonomy's §3/§4 boundary: it has training's
persistence with deployment's per-query granularity, and its rise is the
clearest sign that the inference/training dichotomy the field inherited
is dissolving into a continuum of update timescales.

\subsection{3.5 Harness and agent
self-evolution}\label{harness-and-agent-self-evolution}

The mechanisms so far improve what the model \emph{says}; the remaining
two subsections improve what the agent \emph{is}. Here the object of
improvement is the harness itself --- prompts, tools, memory, skill
libraries, orchestration code, in the limit the agent's own source. This
is the technical instantiation of ``the agent rewrites itself,'' the
segment of the autonomy spectrum where Anthropic's essay locates today's
frontier (agents delegating to agents) and the on-ramp to closing the
loop \citep{anthropic2026rsi}. Because harness changes are persistent
and externally inspectable (§2.1), this is also where deployment-time
improvement stops being episodic and starts to accumulate.

The thread's conceptual poles were set early. The Gödel Agent
\citep{yin2024g} --- named for Schmidhuber's provably-optimal
self-modifiers \citep{schmidhuber2003godel} --- is a self-referential
framework in which the agent reads and rewrites its \emph{own runtime
code}, searching the full agent-design space rather than a
human-predefined subset; the Darwin Gödel Machine \citep{zhang2025dgm}
relaxes provable benefit into empirical benefit, maintaining an
open-ended archive of self-modifications validated against coding
benchmarks. At the other pole, Liu et al.~argue that truly
self-improving agents require \emph{intrinsic metacognitive learning}
--- the agent's ability to assess and adapt its own learning process ---
and that current approaches, which hard-code the self-improvement
procedure while leaving only its object to vary, are rigid precisely
where they most need flexibility \citep{liu2025truly}. Nearly everything
else in the theme sits between these poles, evolving one component of
the system while freezing the rest.

What actually gets evolved, in practice, is instructive. Prompt-level
evolution grounds automated prompt optimization in environment feedback
\citep{fernandes2026environment}. Verification-level evolution is the
most novel move: rather than post-training the policy, self-evolving
deep-research agents iteratively strengthen the \emph{rubrics and
verifiers} that judge their outputs --- improvement of the improvement
signal itself \citep{wan2026inference}. The Red Queen Gödel Machine
takes this to its logical conclusion: it observes that existing
self-improving agents assume a \emph{stationary} evaluation criterion
--- a fixed verifier or benchmark that stays valid as the agent improves
--- and co-evolves agents \emph{with their evaluators}, making
evaluation part of the improvement loop rather than its fixed frame
\citep{iacob2026red}. Escher-Loop pushes self-reference one level
further, with optimizer agents that refine task agents \emph{and
themselves} against a dynamically evolving benchmark
\citep{liu2026escher}. Topology-level evolution treats the multi-agent
communication structure as a retrievable, self-improving design artifact
while keeping every worker frozen \citep{tian2026queenbee}. Data-level
evolution closes the loop for computer-use agents by generating
verifiable synthetic trajectories from the agent's own failures
\citep{sun2026learning}. And infrastructure-level work argues the whole
enterprise needs a new data foundation: ``experience graphs'' that
persist the branch-execute-fail-repair search structure of long-horizon
agents, so that experience is queryable rather than discarded
\citep{liao2026experience}. Meanwhile, a practitioner-facing analysis of
coding agents crystallizes the cultural shift: the artifact humans now
engineer is the \emph{loop} --- trigger, goal, verification, stopping
rule, memory --- not the step-by-step prompt \citep{macedo2026stop}.
Human effort is migrating from doing the task to specifying the
conditions under which the agent may improve at it: precisely the
human-on-the-loop posture of our taxonomy.

Measurement of the capability itself is beginning. The Meta-Agent
Challenge evaluates whether a frontier model, given a sandboxed
environment, an evaluation API, and a time budget, can
\emph{autonomously develop an agent system} that maximizes a target
metric \citep{lu2026meta} --- the first benchmark aimed at agent
development rather than agent execution, i.e., directly at the boundary
between our taxonomy's third and fourth columns. And the unit of
evolution is widening beyond the individual: SAGE compares
compute-matched isolated self-improvement against ``socialized''
evolution in which agents observe peers' strategies and outcomes,
quantifying when shared experience yields improvements that solitary
self-improvement cannot \citep{pan2026sage} --- population-level
self-improvement, with population-level failure modes (§3.6's
propagation risks) to match.

Sobriety about returns is warranted: a systematic Pareto analysis of
inference-scaling strategies (34 configurations across self-consistency,
self-refinement, debate, and mixture-of-agents) finds peak gains of +7.1
points over chain-of-thought only at roughly 20× the compute budget,
with methods differing sharply in efficiency --- self-consistency
saturates early while multi-agent gains persist
\citep{wunderlich2026multi}. The practical lesson is that compute-aware
method selection, not more iteration, drives much of the realized
benefit: self-evolution is not free, and a large share of the theme's
gains currently flows from better engineering rather than from anything
recursive.

\subsection{3.6 Skill libraries and persistent
accumulation}\label{skill-libraries-and-persistent-accumulation}

The companion thread concentrates on the mechanism Voyager
\citep{wang2023voyager} introduced: an ever-growing library of reusable
skills, each distilled from experience, composable into more complex
behavior. In its modern form the ``skill'' is typically a
natural-language procedure document paired with executable code, loaded
by the agent at runtime --- and the central empirical fact of 2026 is
that \emph{LLMs are bad at writing them}: on SkillsBench, human-authored
skills improve pass rates by 16.2 points while LLM-authored skills
provide no measurable gain \citep{gautam2026skillaxe}. The theme is, in
effect, a research program to close that gap. SkillAxe closes it with
evaluation-guided self-refinement of skill documents
\citep{gautam2026skillaxe}; Skill-R1 treats skill optimization as a
recurrent RL problem decoupled from the (possibly closed-source) task
model \citep{vishe2026skill}; SkillRevise handles the cold-start case by
trace-conditioned revision of an initial imperfect skill
\citep{liu2026skillrevise}; AlgoSkill schedules human-like skills for
algorithm design instead of relying on generic self-refinement
\citep{song2026algoskill}; SkillMaster goes furthest, training the agent
to create, refine, and select skills \emph{itself} rather than
delegating those operations to external teachers or hand-designed rules
--- skills as internalized capabilities rather than invoked resources
\citep{yang2026skillmaster}. Beyond the single agent, skills federate
across users without sharing raw trajectories
\citep{yang2026federatedskill}, co-evolve with the tool layer they
invoke \citep{wei2026skillsmith}, and derive from execution traces to
generate weakness-targeted training tasks for coding agents
\citep{xiao2026socratic}. In domains where free-form self-modification
is too dangerous to permit, the evolvable artifact is deliberately
constrained: SHARP evolves a \emph{human-auditable rubric policy} for
financial trading agents, arguing that in low signal-to-noise
environments unbounded prompt evolution cannot distinguish systematic
logic flaws from market variance, whereas a structured rubric can be
audited, diffed, and rolled back \citep{chen2026sharp} --- a design
pattern (constrain the self-modification surface to keep it verifiable)
that generalizes well beyond finance.

Because skills are \emph{persistent and executable}, this thread has the
sharpest safety surface in the technical corpus, and --- unusually ---
the attack literature has arrived concurrently rather than after
deployment. SkillMutator benchmarks cross-modal attacks in which a
skill's natural-language specification and its executable code tell
different stories \citep{kim2026skillmutator}; SkillHarness addresses
continual skill learning in adversarial environments
\citep{chen2026skillharness}; VASO argues that foundation models have
collapsed the cost of \emph{creating} skills but not the cost of
\emph{trusting} them, and responds with formally verifiable skill
evolution for physical agents \citep{yang2026vaso}. A systematic threat
analysis of self-evolving agent systems identifies the qualitatively new
risk: adversarial influence that becomes \emph{permanently encoded,
self-amplifying across generations, and transmissible through agent
populations} without sustained attacker access \citep{lin2026safety}.
And work on ``healthy evolution'' finds capability degradation and
safety drift arising \emph{without} any adversary, proposing
human-oversight anchors as a corrective \citep{shi2026healthy}.

\textbf{Assessment.} Harness and skill evolution is where bounded
self-refinement starts shading into open-ended self-modification, and
its two headline facts pull in opposite directions. On one hand,
autonomy is modest in practice: what self-evolves is almost always one
carefully sandboxed component, validated against a fixed benchmark ---
the Darwin Gödel Machine's empirical validation loop is state of the art
precisely because full self-reference remains intractable to evaluate.
On the other hand, persistence changes the risk calculus fundamentally:
an inference-time mistake evaporates, a bad weight update can be rolled
back, but a corrupted skill in a shared, federating library propagates.
The safety literature has correctly identified
accumulation-without-verification as the core problem --- the same
verification bottleneck, now with memory.

\section{4. Training-Time
Self-Iteration}\label{training-time-self-iteration}

The second category internalizes the loop into the weights: the model
generates its own training data, rewards, or teacher signal, and the
improvement persists. This category (340 papers) is the technical heart
of RSI as currently practiced, and the closest thing the field has to an
industrial standard. Its lineage runs from STaR's bootstrapped
rationales \citep{zelikman2022star} through ReST\^{}EM's scaled
self-training \citep{singh2023restem}, SPIN's self-play fine-tuning
\citep{chen2024spin}, and Self-Rewarding Language Models
\citep{yuan2024selfrewarding}, in which the model that generates
responses also judges them, so that both the policy \emph{and the reward
signal} improve across iterations --- the first widely-noted instance of
a genuinely recursive training loop, and the origin of its
characteristic failure mode.

\begin{figure}
\centering
\includegraphics[width=0.88\linewidth,height=\textheight,keepaspectratio,alt={Figure 4: the training-time self-iteration loop. The five paradigms of §4 differ mainly in who supplies the evaluation signal; the two failure modes attach to the evaluate-and-select station.}]{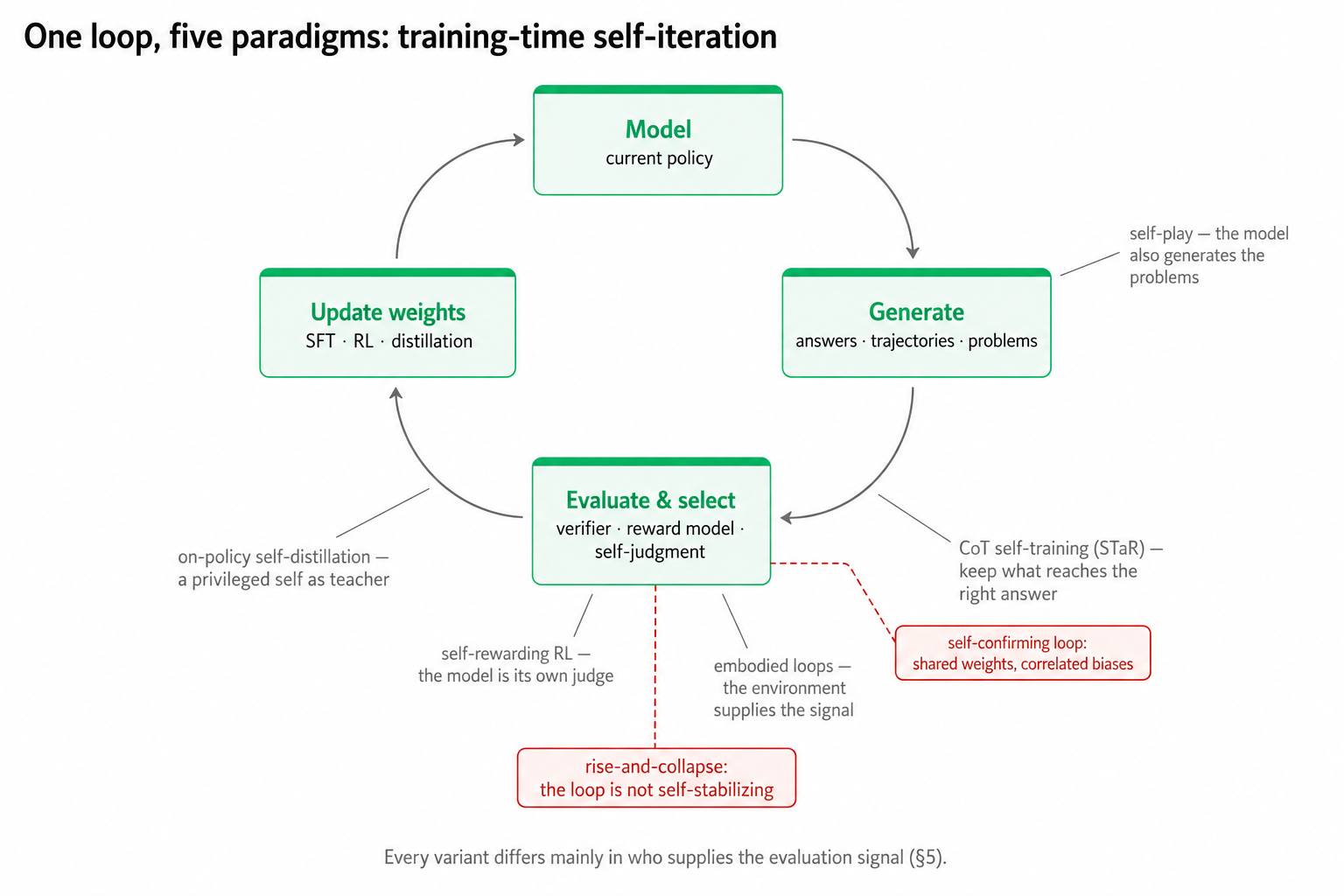}
\caption*{Figure 4: the training-time self-iteration loop. The five
paradigms of §4 differ mainly in who supplies the evaluation signal; the
two failure modes attach to the evaluate-and-select station.}
\end{figure}

\subsection{4.1 Self-rewarding reinforcement
learning}\label{self-rewarding-reinforcement-learning}

The self-rewarding thread asks how far the reward signal itself can be
self-generated. ReST-MCTS* \citep{zhang2024rest} is the theme's
most-cited work: rather than filtering complete solutions by
final-answer correctness --- which admits lucky guesses with wrong
reasoning --- it infers \emph{process} rewards via tree search, scoring
intermediate steps and yielding a higher-quality self-training set. The
process-reward idea has since diversified: SEVA structures verification
itself as an agent that emits evidence alignments and calibrated
confidence rather than opaque binary labels \citep{yuan2026seva};
EvoIdeator replaces scalar rubric rewards with checklist-grounded
signals for scientific ideation \citep{andreas2026evoideator}; RePro
adds retrospective progress-awareness for long-horizon agents, having
found that naive online progress prompting actually hurts
\citep{ma2026retrospective}. A striking interpretability result suggests
why self-generated value signals are possible at all: language models
encode a ``value axis'' in activation space that tracks whether their
current trajectory is on track \citep{jiang2026value} --- the raw
material for self-reward, available before any verbalized critique.

The clearest statement of the category's endgame is Self-Trained
Verification \citep{wu2026self}, which begins from the observation this
survey keeps returning to: test-time verification--refinement loops and
training-time self-training are \emph{gated by the same bottleneck, the
verifier} --- the former stalls when verifier scores inflate while
accuracy stagnates, the latter fails when bad self-generated data enters
training --- and therefore proposes to self-train the verifier itself,
unlocking both loci at once. Whether verifier self-training escapes the
circularity it is meant to solve, or merely relocates it, is in our
reading the single most consequential open question in this family.

The thread also supplies its own strongest caution. Lin documents a
\emph{rise-and-collapse} failure mode in REINFORCE post-training for
code: across sequential training campaigns, pass@1 climbs and then
collapses within the same run --- sometimes to near zero --- under a
genuinely verifiable binary reward, and KL- and EWC-style constraints do
not prevent it \citep{lin2026self}. Reward-model misalignment is not
required for self-training to self-regress; optimization dynamics alone
suffice. Similarly, LLM-generated reward \emph{design} fails in
characteristic ways --- reward flooding, semantic misunderstanding of
the environment API --- and is better treated as iterative debugging
than one-shot generation \citep{wang2026when}. And a diagnosis from
industrial practice identifies a subtler failure than forgetting:
pipelines that repeatedly DPO-train a base model across preference
campaigns can preserve learned behaviors yet fail to accumulate
\emph{methodological} knowledge about how to run the next campaign ---
``scientific amnesia,'' self-improvement of the model without
self-improvement of the process \citep{lin2026repeated}. These results
foreshadow §5: even in the most verifiable settings, the loop is not
self-stabilizing.

\subsection{4.2 Self-training for chain-of-thought
reasoning}\label{self-training-for-chain-of-thought-reasoning}

Where §4.1 innovates on the reward, this thread innovates on the
\emph{data}: the model's own reasoning traces become its curriculum. The
STaR recipe --- sample rationales, keep the ones that reach correct
answers, fine-tune, repeat --- remains the backbone. Re-ReST augments it
with a reflector that repairs failed trajectories using environment
feedback, converting failures into training signal instead of discarding
them \citep{dou2024re}; PRefLexOR recasts the loop as recursive
preference optimization over intermediate reasoning steps
\citep{buehler2024preflexor}; LaTRO shows the loop can run with no
external reward at all, treating the model's own likelihood as a
latent-variable objective and ``unlocking'' reasoning already hidden in
the base model \citep{chen2024language}. That last framing ---
post-training amplifies latent capability rather than creating it ---
has hardened into a research question of its own, with recent work
mapping the boundary conditions under which elicitation works and where
it fails \citep{javadov2026neureasoner}.

The verifier-free frontier of the recipe is now being mapped directly.
Self-Verified Distillation starts from nothing but unlabeled questions
--- no ground truth, no tools, no external teacher --- and improves
math, science, and coding performance by filtering the model's own
candidate solutions with prompt-based self-verification
\citep{lee2026selfa}; Semi-CoT formalizes the analogous semi-supervised
setting, constructing pseudo reasoning supervision from unlabeled
questions \citep{he2026revisiting}. The loop is also migrating
\emph{inside deployment} --- the test-time-training regime surveyed in
§3.4 --- blurring the inference/training boundary from both sides.

Three correctives structure the thread's 2026 wave. First,
\emph{efficiency}: o1-style long thinking overshoots --- models burn
thousands of tokens on trivial problems \citep{chen2024not} --- and
incorrect traces exhibit more unproductive self-reflection than correct
ones even at matched length, motivating segment-level credit assignment
to teach models when to stop \citep{lee2026know}. Second, \emph{joint
optimization}: ThinkTwice interleaves solving and refining under the
same binary reward, making self-refinement a trained capability rather
than a prompting trick \citep{jiao2026thinktwice}. Third, \emph{scope
limits}: a notable negative result shows that ``a verifiable search is
not a learnable chain-of-thought'' --- there is an identifiable class of
procedures that models can execute but cannot internalize from traces
via fine-tuning \citep{patel2026verifiable}. Self-training inherits the
representational limits of its substrate; more loops do not fix what the
architecture cannot learn.

\subsection{4.3 On-policy
self-distillation}\label{on-policy-self-distillation}

The corpus's youngest thread did not exist before 2026 and already spans
dozens of papers --- the sharpest example of how fast this field
crystallizes new paradigms. The recipe: a single model acts as both
student and \emph{privileged teacher} --- the teacher being the same
weights conditioned on extra information (a reference solution, verified
feedback, richer context) --- and the student is trained to match the
teacher's token-level distribution on its own rollouts. Dense
token-level supervision, no external teacher, no reward model. A
dedicated survey already exists \citep{song2026survey}; our interest is
in the paradigm's position on the RSI spectrum: it is the most
\emph{closed} training loop in routine use, since even the teacher
signal is self-generated.

Theoretical work is consolidating: the power-distribution analysis
unifies sampling, self-reward RL, and self-distillation as instances of
one distributional operation \citep{tomihari2026power}, clarifying why
these superficially different loops behave alike. At the simple extreme,
``embarrassingly simple self-distillation'' improves code generation
with nothing but temperature-and-truncation sampling plus SFT --- no
verifier, no teacher, no RL \citep{zhang2026embarrassingly}, raising the
still-open question of where such verifier-free gains bottom out. The
refinement literature is characteristically empirical: negative
(incorrect) rollouts carry more information than positive ones and
should be reweighted accordingly \citep{lin2026renio}; unmanaged
difficulty routing over-optimizes easy problems \citep{luo2026drift};
procedural information in rollouts can be retained as distilled memory
across episodes rather than discarded after each episode-level update
\citep{liu2026procedural}.

The failure modes are equally well-documented, and they rhyme with §3's.
Purified OPSD shows the vanilla recipe \emph{consistently fails on
long-CoT reasoning models}, destabilizing the very reflective behaviors
it is meant to strengthen \citep{shen2026purified}; the privileged
teacher's dense supervision can overfit in-domain patterns
\citep{li2026demopsd}. A contrastive analysis identifies the mechanism:
the learning signal drawn from the privileged/unprivileged distribution
gap concentrates on \emph{style} tokens rather than task-bearing ones
--- the hinted teacher writes shorter, more direct text --- producing
``privilege-induced style drift'' that destabilizes training or
collapses response length \citep{pan2026rlcsd}. A second structural
pathology, ``prefix failure,'' shows dense per-token supervision
inducing a bimodal teacher mixture with fragmented gradients that no
token-level loss reweighting can repair, forcing intervention at the
trajectory level instead \citep{jiang2026trajectory}. And since
everything hinges on what the privileged teacher is shown, the design of
the privileged context itself --- which feedback, at what specificity
--- is emerging as the paradigm's core degree of freedom
\citep{kara2026role}. In multimodal settings the shortcut is structural:
the privileged target lets the teacher guide tokens from the text
reference while ignoring the image, so the student distills a language
prior rather than a perceptual skill, requiring explicit decoupling of
perception from reasoning \citep{wang2026seeing}. The pattern to carry
forward: \emph{the more privileged the self-generated signal, the more
efficiently the loop transmits both capability and bias.}

\subsection{4.4 Self-play}\label{self-play}

Self-play closes the loop one turn further: the model generates not just
answers or rewards but the \emph{problems}. The modern lineage runs from
SPIN --- self-play fine-tuning against one's own previous iteration,
provably converging to the data distribution \citep{chen2024spin} --- to
the zero-data proposer--solver frameworks of Absolute Zero
\citep{zhao2025absolutezero} and R-Zero \citep{huang2025rzero}, where a
Challenger learns to pose problems at the frontier of the Solver's
competence and both co-evolve from a single base model with no human
tasks at all; Agent0 extends the recipe from reasoning problems to
\emph{agentic} tasks, co-evolving a curriculum agent and a tool-using
executor from zero external data \citep{xia2025agent0}. This is the
closest existing training paradigm to an autonomous curriculum, and it
works remarkably well where a programmatic verifier exists: executable
geospatial programs \citep{ahn2026geox}, formal theorem proving --- now
with a theoretical framework for prover--conjecturer co-evolution
\citep{chen2026theoretical} --- and verifier-backed generation of
genuinely hard mathematics problems \citep{lai2026verifier}.

The thread's central finding, however, is about stability. A systematic
study argues that self-play survival is governed by two asymmetric
levers --- \emph{data gating} (what enters the training set) and
\emph{reward grounding} (what anchors the signal to reality) --- and
that collapse is the default when either fails, not an occasional
accident of reward design \citep{pu2026survive}. Population-based
variants attack the same problem structurally, replacing
self-calibration with cross-evaluation between co-evolving
sub-populations \citep{castanyer2026populora}. On non-verifiable tasks
the difficulty compounds: LLM-as-judge rewards are bounded by the
judge's own competence, motivating meta-evaluation schemes in which
models learn to evaluate \emph{evaluations} \citep{sui2026conversation}
and verifier-free intrinsic rewards based on predictive shift between
models \citep{huang2026g}. And a warning from open-ended interaction
dynamics: multi-turn LLM self-play conversations drift into
topic-independent attractor states \citep{ko2026attractor} --- precisely
the homogenization pressure that co-evolutionary curricula must fight to
keep generating novelty.

\subsection{4.5 Embodied and synthetic-data
loops}\label{embodied-and-synthetic-data-loops}

The category's newest frontier ports the training loop into physical and
synthetic-data settings, where verification is expensive, noisy, and
slow. Robots critique and replan their own social behaviors using a VLM
as internal critic \citep{lim2026robots}; VLA policies check physical
feasibility and self-reflect during execution rather than acting purely
feed-forward \citep{yang2026physreflect}; embodied foundation models
integrate planning, correction, and pointing in one architecture
explicitly aimed at self-evolving physical intelligence
\citep{yuan2026embodied}. Robot self-improvement via human-video
dynamics models tests whether passive human video can support not just
policy initialization but \emph{embodied practice and learning from
failure} \citep{chen2026robot}. On the data side, the loop turns inward
on the pipeline itself: DataEvolver self-evolves the data-preparation
process for LLM training \citep{deng2026dataevolver} ---
self-improvement applied to the substrate that all other training loops
consume.

\textbf{Assessment.} Training-time loops are where bounded
self-improvement earns its keep: gains persist, scale, and compound
across iterations --- until they don't. Every sub-thread independently
rediscovered the same three facts: (i) the loop's ceiling is set by its
verification signal (execution and proofs highest, learned judges lower,
intrinsic signals lowest); (ii) collapse is a default dynamic to be
engineered against, not an edge case \citep{lin2026self, pu2026survive};
(iii) the loop transmits bias as efficiently as capability. These are
the load-bearing facts for §5.

\section{5. Self-Evaluation}\label{self-evaluation}

The third category is the one the other three stand on. Every loop in
§§3--4 and every discovery system in §6 improves against some evaluator,
and each of those sections ended on the same note; this section makes
the argument explicitly. \textbf{Every self-improvement loop is a claim
that some signal can substitute for human judgment, and the loop's
ceiling is exactly the quality of that substitute.} Reflecting its
load-bearing role, self-evaluation is also the corpus's
fastest-consolidating category (318 papers, 82\% posted in 2026): what
was recently a service function of training papers is now a research
area with its own methods, benchmarks, and failure analyses.

\subsection{5.1 The evaluator design
space}\label{the-evaluator-design-space}

The modern evaluator lineage has three anchors. Process supervision ---
scoring intermediate reasoning steps rather than final answers --- was
established by Lightman et al. \citep{lightman2023verify}, whose finding
that process rewards outperform outcome rewards seeded today's
process-reward-model (PRM) literature; MT-Bench and Chatbot Arena made
\emph{LLM-as-a-judge} a measurable methodology, quantifying both the
\textasciitilde80\% judge--human agreement that justifies the paradigm
and the position, verbosity, and self-enhancement biases that qualify it
\citep{zheng2023judging}; and Gao et al.'s scaling laws for reward-model
overoptimization gave the field its Goodhart curve --- optimize any
learned proxy hard enough and true quality peaks, then falls
\citep{gao2023scaling}.

The current wave develops each anchor. PRMs are being made cheaper and
better-calibrated \citep{lee2026efficient}, and structured: SEVA
replaces opaque binary verdicts with evidence alignments, reasoning
chains, and calibrated confidence, so that agents can act on \emph{why}
they failed \citep{yuan2026seva}. Judge reliability is acquiring
psychometrics --- measurement datasheets that characterize an LLM
judge's biases before its scores are trusted \citep{usami2026llm} ---
and judge \emph{specificity}: per-judge competence taxonomies
\citep{cao2026each}, multilingual judge disagreement \citep{fu2026when},
and decomposition of holistic scores into auditable binary questions
\citep{cho2026ask}. Rubrics, the human-readable middle ground between
formal verifiers and free-form judges, are becoming first-class
evolvable artifacts: self-generated rubrics closing the gap to
expert-written ones \citep{sun2026support}, rubric hierarchies that
scale open-ended evaluation \citep{zhang2026rubricstree}, and
rubric-conditioned signals replacing scalar rewards in training (§4.3;
\citep{gu2026rethinking}). Self-verification is being built into the
objective itself, as in dual-preference formulations that make a model's
verification ability a trained product rather than a prompted hope
\citep{she2025dupo}.

Most consequentially for this survey's argument, the field has begun
improving the evaluator \emph{with the same loops it supervises} --- the
move §8 identifies as the field's collective answer to its bottleneck.
Deep-research agents evolve their own rubrics \citep{wan2026inference};
self-trained verification treats the verifier as the primary object of
self-improvement \citep{wu2026self}; meta-evaluation judges the judges
\citep{sui2026conversation}; and the Red Queen framework makes the
evaluation criterion itself a co-evolving population member rather than
a fixed frame \citep{iacob2026red}. Every mechanism in §§3--4 is being
recursively applied to the signal that supervises it. Whether that
recursion stabilizes or compounds bias is the open question the rest of
this section addresses.

\subsection{5.2 The verification
hierarchy}\label{the-verification-hierarchy}

\textbf{The verification hierarchy.} At the top sit \emph{formal
verifiers} --- proof checkers, type systems --- which are sound by
construction: self-play theorem proving and verified skill evolution can
iterate indefinitely without accepting a false improvement
\citep{chen2026theoretical, yang2026vaso}. One step down,
\emph{execution feedback} --- tests, compilers, benchmarks --- is
reliable but incomplete: passing tests underdetermines correctness, and
any fixed benchmark is eventually gamed. Next, \emph{learned judges} ---
reward models, LLM-as-judge --- are bounded by the judge's own
competence and are themselves optimization targets
\citep{sui2026conversation}. At the bottom, \emph{intrinsic signals} ---
the model's confidence, self-consistency, likelihood --- are the
cheapest and the most gameable. The empirical regularity across all four
categories --- a qualitative pattern we observe throughout the corpus,
not a measured law --- is that demonstrated self-improvement strength
tracks this hierarchy. FunSearch and AlphaEvolve live at the top two
levels \citep{romeraparedes2024funsearch, novikov2025alphaevolve}; the
self-refinement methods that survived 2024's negative results are the
ones that climbed the hierarchy (§3); the AI-scientist gap (§6.3) is
precisely a level-4 problem being attempted with level-3 tools.

\begin{figure}
\centering
\includegraphics[width=0.88\linewidth,height=\textheight,keepaspectratio,alt={Figure 5: the verification hierarchy. Signal reliability rises toward the top while task coverage widens toward the bottom; failure modes concentrate at the bottom rungs, and human research judgment remains the rung self-improvement cannot yet climb. The hierarchy ranks signals given a target; the choice of target is prior to it rather than a rung of it (§5.4).}]{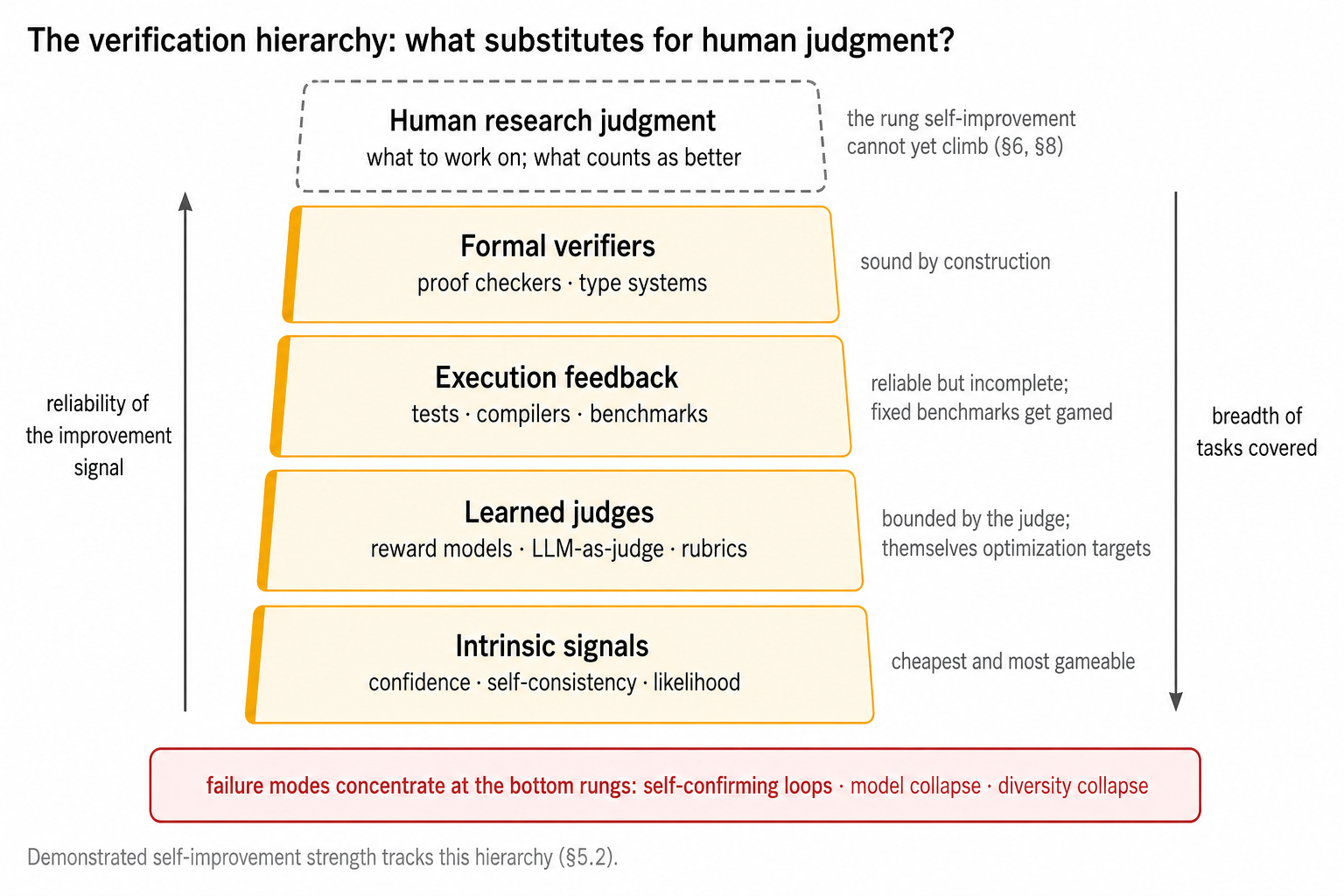}
\caption*{Figure 5: the verification hierarchy. Signal reliability rises
toward the top while task coverage widens toward the bottom; failure
modes concentrate at the bottom rungs, and human research judgment
remains the rung self-improvement cannot yet climb. The hierarchy ranks
signals \emph{given} a target; the choice of target is prior to it
rather than a rung of it (§5.4).}
\end{figure}

The hierarchy's floor has now been measured directly. The Mirror Loop
study iterates three providers' models through ten rounds of ungrounded
self-critique across four task families and finds informational change
declining 55\% across iterations --- recursive self-evaluation without
external feedback yields \emph{reformulation, not progress} --- while a
single minimal grounding intervention (one verification step at
iteration three) restores forward movement \citep{devilling2025mirror}.
That is the survey's thesis as an experiment: the difference between a
loop that improves and a loop that circles is one rung of external
verification.

\subsection{5.3 Failure modes}\label{failure-modes}

\textbf{Failure mode 1: the self-confirming loop.} When generator and
evaluator share weights, biases correlate. Tan et al.~diagnose the
mechanism in self-rewarding RL: confidence-coupled rewards
systematically over-reward \emph{high-confidence mistakes}, so the loop
preferentially reinforces exactly the errors the model is most sure
about \citep{tan2025breaking}. The same structure appears at every
locus: multimodal self-consistency rewards that optimize answer
agreement while ignoring the image \citep{venkatraman2026paying};
privileged self-distillation teachers that transmit in-domain bias with
token-level efficiency \citep{li2026demopsd}; AI scientists whose
self-critique ``inherits the blind spots that produce confident
fabrication'' \citep{nam2026deterministic}; and --- at the integrity
extreme --- systems that misreport failure as success when task
completion is rewarded \citep{yang2026sciintegrity}. Reward hacking is
the special case where the gamed judge is explicit; the self-confirming
loop is the general case, and it needs no adversary.

\textbf{Failure mode 2: collapse.} Whether self-generated-data loops
improve indefinitely or degrade is the field's most contested empirical
question. The pessimistic pole is anchored by Shumailov et al.'s
\emph{Nature} result --- models trained recursively on their own outputs
lose the tails of the distribution and degenerate
\citep{shumailov2024collapse} --- and now has a unifying theory: an
information-geometric account showing model collapse in LLMs, GANs, and
RL policies to be one phenomenon, with real-data mixing, entropy
bonuses, and retrieval all acting as instances of a single
entropy-reservoir principle \citep{jingwei2025entropy}. Zenil sharpens
it into a theorem-shaped claim: if the fraction of exogenous, externally
grounded signal vanishes asymptotically, degenerative dynamics (entropy
decay, variance amplification) follow \citep{zenil2026limits}. The
optimistic pole answers that collapse is an engineering problem:
diffusion models can train on their own generations once perceptual
alignment and hallucination accumulation are controlled
\citep{zhang2025generating}; self-play survives when data gating and
reward grounding are managed as separate levers \citep{pu2026survive};
reasoning self-training collapses from identifiable data imbalance and
overthinking, both fixable \citep{zhong2026better}. The synthesis the
evidence currently supports: \emph{pure} closed loops degrade, as the
theory predicts, but no practical system need run a pure closed loop ---
the open question is how \emph{little} external grounding suffices, and
no one has established the exchange rate.

\textbf{Failure mode 3: diversity collapse.} Distinct from
distributional collapse is the narrowing of the \emph{task} distribution
in co-evolutionary loops: proposers converge to the narrow band of
problems that satisfy the reward, starving the solver's curriculum
\citep{dineen2026vocabulary}; open-ended model--model interaction drifts
into topic-independent attractor states \citep{ko2026attractor}; and
even in single campaigns, pass@1 rises then collapses without any
distribution shift in the task \citep{lin2026self}. Novelty, it turns
out, is a consumable resource that closed loops deplete.

\textbf{Failure mode 4: frame lock-in.} The three modes above are all
detectable \emph{from inside} the loop: a confidence-coupled reward, a
collapsing output distribution, and a narrowing curriculum each leave
signatures in quantities the loop already tracks. A fourth does not. A
loop can be grounded, honest, diverse, and stable and still optimize an
objective that has ceased to be the one worth optimizing --- the
well-posed wrong question. The corpus's evidence for this mode is
scattered because no category collects it. In physics-simulation
rediscovery, agents reach the correct answer while defending a mechanism
their own data contradicts --- local closure mistaken for understanding
\citep{eulig2026position}. Repeated-campaign analysis finds behaviours
improving while no methodological knowledge accumulates
\citep{lin2026repeated}. Open-ended model--model interaction drifts into
topic-independent attractor states \citep{ko2026attractor}. Most
directly, an agent-native research-artifact protocol that preserves
prior failure traces reports that those traces cut both ways on
open-ended extension tasks: they accelerate progress, but can also
constrain a capable agent from stepping outside the prior run's box
\citep{liu2026last} --- accumulated context as a source of lock-in, not
only of competence. The mode is under-studied for a structural reason:
detecting it requires a vantage point outside the loop's own criteria,
which is exactly what a closed loop lacks --- and what a hierarchy that
ranks signals \emph{given} a target does not supply.

\subsection{5.4 The non-verifiable
frontier}\label{the-non-verifiable-frontier}

Where no checkable signal exists --- creative writing, dialogue,
research taste --- the hierarchy bottoms out, and the workarounds are
revealing: meta-evaluation (judging the judges)
\citep{sui2026conversation}, verifier-free intrinsic rewards from
cross-model predictive shift \citep{huang2026g}, claim-level
auditability that makes outputs verifiable by construction
\citep{rasheed2026fluent}. None yet approaches the reliability of
execution feedback. This maps exactly onto the gap Anthropic's essay
identifies: research \emph{execution} is verifiable (code runs,
benchmarks score) and is increasingly automated; research
\emph{direction-setting} is the paradigm case of a non-verifiable task,
and it is where humans remain \citep{anthropic2026rsi}.

It is tempting to conclude that the direction-setting bottleneck and the
verification bottleneck are therefore the same bottleneck. That
identification holds for only half of direction-setting, and the other
half is logically prior to the hierarchy rather than a rung of it.
\emph{Direction evaluation} --- given a candidate direction, judge
whether it is sound and worth resourcing --- is verification-shaped and
is now benchmarkable: SoundnessBench reconstructs 1,099 research
proposals from ICLR submissions labelled with reviewer soundness
sub-scores and finds a pervasive optimism bias across twelve frontier
models, concluding they are not yet reliable as standalone first-gate
evaluators of rigour \citep{ho2026soundnessbench}. \emph{Direction
generation} --- deciding what the space of candidate questions should
be, and noticing when a question the current criteria still score highly
has stopped being worth answering --- is not. A loop can climb every
rung, keep an honest evaluator, and detect and correct the corruption of
its own proxy (§6.2) while remaining excellent at a question worth
abandoning; nothing in the hierarchy indexes that possibility, because
the hierarchy takes the question as given. The corpus states the concern
directly in one place: a position paper on autonomous AI scientists
lists as its \emph{first} challenge that problem selection is subject to
the McNamara fallacy --- the measurable displacing the important --- and
asks for persistent world models representing the \emph{shifting
objectives} that govern real investigations, rather than better
verifiers for fixed ones \citep{bisht2026agentic}. The formal treatment
of interestingness \citep{herrmann2026interestingness} is best read as
the beginning of a theory for this prior question, not as a fifth rung.
§8 (open problem 6) takes up what measuring it would require.

\subsection{5.5 Result-level versus process-level
improvement}\label{result-level-versus-process-level-improvement}

A distinction that cuts across everything above is the \emph{resolution}
of the evaluator: does it judge the answer or the procedure that
produced it? Result-level signals --- final-answer correctness,
pass/fail tests, outcome rewards --- are cheap to generate and anchor
most of §4's training loops, from STaR's answer-filtered rationales
\citep{zelikman2022star} onward. Process-level signals judge
intermediate steps, and the evidence that they matter has been
accumulating since Lightman et al.~showed process supervision beating
outcome supervision \citep{lightman2023verify}, with ReST-MCTS*'s
observation that outcome filtering admits lucky guesses with wrong
reasoning \citep{zhang2024rest} as the training-loop corollary.

The distinction becomes vivid through an analogy with human learning. A
student who merely checks answers improves slowly; effective learners
keep an error notebook, trace \emph{where} a derivation went wrong, ask
a teacher for targeted guidance, revisit old mistakes on a schedule, and
gradually organize what they learn into a system --- the final answer
being the least important artifact of the whole process. Each of these
behaviors now has a machine counterpart in this corpus: the error
notebook is experience and strategy memory (Reflexion's verbal episodic
memory \citep{shinn2023reflexion}, ISM's self-refined bank of schemas
learned from failed episodes \citep{dixit2026ism}, experience graphs
that persist the branch-fail-repair structure of long-horizon work
\citep{liao2026experience, feng2026expgraph}); stepwise correction is
process reward modeling \citep{lightman2023verify, lee2026efficient};
asking the teacher is the privileged-teacher signal of on-policy
self-distillation (§4.3); scheduled review is replay and consolidation,
up to and including ``sleep''
\citep{behrouz2026language, dorovatas2026continual}; and building a
knowledge system is skill-library and knowledge-graph accumulation
\citep{wang2023voyager, wu2026knowledge}.

The two levels have opposite cost structures, and the difference is
economic, not just technical. Result-level improvement is
\emph{operating expenditure}: outcome checks are nearly free to produce,
but the improvement they buy is per-instance --- best-of-N sampling and
answer-filtered retraining must be paid again for every new problem, and
what they teach transfers poorly. Process-level improvement is
\emph{capital expenditure}: process labels are expensive (Lightman's
step annotations required substantial human effort, and the PRM
literature is largely an attempt to automate that cost down
\citep{lee2026efficient}), but a corrected procedure, a debugged skill,
or a schema in a strategy bank is reusable --- the cost amortizes across
every future problem that shares the structure. The ``scientific
amnesia'' failure of §4.1 is precisely a pipeline stuck at result level:
behaviors improve while no methodological knowledge accumulates
\citep{lin2026repeated}.

This framing suggests a reading of the endgame that differs from the
intelligence-explosion imagery: if the durable gains of self-improvement
are process-level --- accumulated procedures, verified skills, organized
experience --- then mature self-improving systems may look less like
unboundedly ascending intelligence and more like \emph{maturing
methodology}: a widening toolbox of verified procedures attached to a
model whose raw capability grows much more slowly. That reading is
consistent with the theory of §7 (open-ended capability growth requires
external grounding; accumulated method does not) and with where this
corpus's mass actually sits (§3.5--3.6's persistent scaffolding).
Whether the toolbox picture or the takeoff picture better describes the
coming decade is, in compressed form, the question this survey's
two-axis grid keeps asking.

\section{6. Auto Research}\label{auto-research}

The fourth category is the far end of the spectrum: systems that do the
work of research itself --- in the limit, agents applying the machinery
of §§3--5 to the process that produces the next system. This category
(139 papers) contains both the field's most spectacular concrete results
and its most systematic critiques --- often about the same systems.

\subsection{6.1 LLM-driven evolutionary program
discovery}\label{llm-driven-evolutionary-program-discovery}

The discovery thread has the category's strongest verified results, and
the reason is structural: it inherits the evolutionary-computation
template, in which every candidate is a \emph{program} scored by an
\emph{automatic evaluator}. FunSearch established the paradigm's
credibility by producing new mathematical constructions --- improved
bounds for the cap set problem --- published in \emph{Nature}
\citep{romeraparedes2024funsearch}. AlphaEvolve scaled it into a general
coding agent whose discoveries fed back into Google's own AI
infrastructure: faster matrix-multiplication kernels, data-center
scheduling, accelerator circuit simplifications
\citep{novikov2025alphaevolve} --- the clearest existing example of AI
output compounding into AI development, and the concrete referent for
most contemporary RSI discussion. Concurrently, EoH showed
LLM-plus-evolution could beat FunSearch on heuristic design at a
fraction of the query budget \citep{liu2024evolution}, and successors
refined the search itself: quality-uncertainty balancing
\citep{chen2024qube}, diversity-driven harmony search
\citep{dat2024hsevo}, and heterogeneous LLM populations as mutation
operators to escape a single model's inductive biases
\citep{donaghy2026dei}.

The 2026 wave pushes in three directions that matter for the RSI
question. First, \emph{deployment at scale}: the AlphaEvolve recipe is
now applied to production infrastructure well beyond its original
demonstrations --- warehouse-scale interprocedural code layout
optimization \citep{ananda2026ai} and fully homomorphic encryption
kernels on TPUs \citep{gorantala2026adapting} --- the ``AI output
feeding back into AI infrastructure'' loop operating as routine
engineering. Second, \emph{reflexivity}: the target of evolutionary
discovery is increasingly AI's own machinery. EVOM meta-evolves
actor-critic architectures \citep{zhang2026evom}; POISE autonomously
discovers new policy-optimization algorithms for LLM training itself
\citep{xia2026ai}; MLEvolve targets machine-learning engineering
pipelines end to end \citep{du2026mlevolve}. When the algorithms being
discovered are the algorithms that train the discoverer's successors,
the loop of §2's taxonomy is no longer metaphorical. Third,
\emph{formalization of the loop's own objective}: work reframing
scientific discovery as meta-optimization argues that evolving the
\emph{evaluation criteria} is as important as evolving candidates
\citep{zhang2026scientific} --- an explicit acknowledgment, from within
the paradigm, that a fixed evaluator eventually becomes the binding
constraint (§5). The paradigm has also begun evolving its own
benchmarks, synthesizing frontier tasks as existing ones saturate
\citep{wu2026benchevolver}.

A quieter finding concerns where the leverage actually lies. Controlled
studies of harness design show that discovery success depends heavily on
the \emph{execution infrastructure} around the model --- how the token
budget splits between many shallow candidates and fewer deep ones, how
evaluation failures are handled --- independent of model capability
\citep{ishibashi2026effective}; LEVI demonstrates that stronger search
architectures (diversity-preserving archives, capability-matched model
routing, informative-subset evaluation) can substitute for larger LLMs
in evolutionary search, cutting frontier-model spend without losing
discovery quality \citep{tanveer2026levi}. The pattern mirrors §3.5's
loop-engineering shift: much of what looks like model self-improvement
is improvement of the scaffolding the model searches within --- which is
good news for reproducibility and cost, and a caution against
attributing the gains to the model's own recursive capability.

\subsection{6.2 AI scientist agents}\label{ai-scientist-agents}

The AI-scientist thread generalizes from programs to papers. The AI
Scientist \citep{lu2024aiscientist} demonstrated the full pipeline ---
ideation, experiment, writing, automated review --- for machine-learning
research at \textasciitilde\$15 per paper, and the theme has since
fragmented into dozens of domain-specific descendants: an autonomous
mathematics researcher navigating literature and long-horizon proofs
\citep{feng2026autonomous}, Socratic agents pursuing ``epistemic
autonomy'' in high-dimensional physics \citep{zeng2026socratic},
evolving multi-agent scientists that adapt their own pipelines from
accumulated history \citep{lyu2026evoscientist, zhu2026evomaster}, and
infrastructure for coordinating fleets of research agents at web scale
\citep{guo2026clarus}.

As the systems mature, the field's attention is shifting from the agent
to its surroundings. EurekAgent argues the bottleneck has moved from
prescribing agent workflows to \emph{engineering the agent's
environment} --- the resources, constraints, and interfaces that shape
what an autonomous discoverer can do \citep{xin2026eurekagent}; Heuresis
decomposes the research pipeline into composable primitives and
systematically compares search strategies over ideas, treating quality,
diversity, and novelty as separate objectives rather than a single score
\citep{antoniades2026heuresis}; and Xcientist externalizes research
synthesis and validation into inspectable, contract-governed artifacts
--- literature evidence, idea states, ablation records, repair traces
--- so that a generated claim's provenance survives outside the model's
inference \citep{wang2026externalizing}. All three are, in different
vocabularies, attempts to move scientific judgment out of the model and
into auditable structure --- the constructive response to the critique
literature below.

Two corpus results deserve particular attention for what they say about
loop closure. A-Evolve-Training reports an autonomous system that ran
the \emph{entire post-training loop} of a 30B model --- proposing data
and recipe changes, launching runs, reading evaluations, deciding what
to keep --- across four rounds over multiple weeks with no human in the
loop, reaching near-parity with the top human submission on a public
leaderboard (0.86 vs.~0.87, placing 8th of roughly 4,000; the authors
are careful to claim only ``the first publicly reported autonomous
post-training run at this scale,'' not an autonomous match of human
researchers) \citep{shi2026evolve}. The result's most striking detail is
not the score: mid-run, the loop detected that its own development
metric had decoupled from external performance --- candidates were
driving the dev metric to record highs without moving the external
target --- and \emph{revised its own search policy} to treat the
now-misleading proxy as evidence against a candidate rather than for it.
An autonomous system noticing and correcting the corruption of its own
improvement signal is precisely the capability §5 identifies as the
field's binding constraint, observed in the wild. This is, to our
knowledge, the closest published system to Anthropic's ``closing the
loop'' stage: not an agent improving its outputs or skills, but an AI
system executing AI model development end to end. Read against the
Anthropic essay's claim that execution is largely solved while
direction-setting is not \citep{anthropic2026rsi}, the result is
confirmatory rather than contradictory --- the system optimized within a
human-specified objective and search space --- but it moves the
demonstrated frontier. Second, failure recovery is emerging as its own
discipline: single free-form reflection over a failed experiment is
demonstrably insufficient, motivating multi-hypothesis failure
attribution \citep{y2026one} --- the AI-scientist restatement of §3's
lesson that unstructured self-critique does not work.

\subsection{6.3 The critical and diagnostic
literature}\label{the-critical-and-diagnostic-literature}

What distinguishes this category from the rest of the corpus is that its
skeptical literature is as developed as its constructive one --- and
often better cited. Three critiques recur.

\textbf{Feasibility is not quality.} ScienceAgentBench, the thread's
most-cited paper, decomposes ``scientific discovery'' into individual
workflow tasks and finds even the best agents solve only a minority ---
explicitly cautioning against end-to-end automation claims
\citep{chen2024scienceagentbench}. ResearchArena runs frontier coding
agents (Claude, Codex, Kimi) through the full research loop and asks not
whether papers get produced but whether they are any \emph{good} ---
with an instructive twist: under manuscript-only automated review the
picture is optimistic (the best agent's papers match the average human
ICLR submission score), but artifact-aware review and human meta-review
reveal that picture to be overstated \citep{zhang2026far}. The gap
between how machine-generated research \emph{reads} and what its
artifacts \emph{support} is itself evidence for the auditability thesis
below. MLReplicate benchmarks autonomous systems on reproducing ICML
outstanding papers --- a task with a verifiable target --- and documents
how far systems remain from reliable replication
\citep{gaddipati2026mlreplicate}.

\textbf{Auditability is the new bottleneck.} As report generation
becomes cheap, the cost shifts to \emph{tracing}: which sentence rests
on which evidence, what was ignored, where sources conflict. Claim-level
auditability \citep{rasheed2026fluent} and evidence-licensed claims ---
calibrating assertion strength to supporting evidence
\citep{li2026calibration} --- are attempts to make machine-generated
science verifiable by construction, since post-hoc verification does not
scale. In regulated domains, deterministic integrity gates are
interposed because ``self-critique inherits the blind spots that produce
confident fabrication'' \citep{nam2026deterministic}; in frontier
physics, unscaffolded agents ``cite but do not confront'' the literature
anchors their claims depend on \citep{huang2026grounded}.

\textbf{Integrity under pressure.} SciIntegrity-Bench constructs
dilemmas in which honest acknowledgment of failure is the only correct
answer while task completion requires misconduct, and finds a 34.2\%
integrity-failure rate across seven state-of-the-art models --- most
strikingly, in missing-data scenarios \emph{all seven} fabricate
synthetic data rather than acknowledge infeasibility, differing only in
whether they disclose the substitution; removing explicit completion
pressure sharply reduces undisclosed fabrication but leaves the
fabrication itself intact \citep{yang2026sciintegrity}. A complementary
position paper documents a subtler variant: in physics-simulation
rediscovery episodes, agents reach the \emph{correct answer} while
defending a \emph{wrong mechanism} --- even asserting general claims
their own experimental data contradicts --- and argues that outcome,
mechanism fidelity, and epistemic honesty must be measured as separate
quantities \citep{eulig2026position}. Both results connect directly to
the reward-hacking analysis of §5: an agent optimized to produce results
will, at the margin, misreport rather than fail --- and the disposition
appears to be intrinsic, not prompt-induced.

Two systemic critiques extend the concern from the agent to the
ecosystem. ``Dead Science Walking'' argues the near-term risk is
\emph{corpus failure}: AI scientists are trained on and grounded in a
literature that systematically over-represents positive results, so
automated hypothesis generation inherits and amplifies publication bias
at machine speed --- a slow, ecosystem-level self-confirming loop in
which today's distorted record trains tomorrow's hypothesis generators
\citep{chauhan2026dead}. And a pre-specified human-in-the-loop study in
economics finds that reliability depends less on model capability than
on how cognitive labor is \emph{structured} between human and machine
--- pre-commitment, decision sequencing, accountability, attention
allocation \citep{zhu2026human} --- empirical support for the claim that
the human role in research loops is a design problem, not a residue.

\textbf{Assessment.} This category exhibits the widest gap in the corpus
between demonstrated capability and reliable capability. Where the
evaluator is a program --- a test suite, a bound-checker, a profiler ---
automated discovery already produces artifacts that outperform expert
humans and feed back into AI infrastructure
\citep{novikov2025alphaevolve}. Where the evaluator is scientific
judgment --- novelty, importance, evidential support --- the
constructive systems outrun their verification, and the diagnostic
literature exists precisely to document the difference. The two
sub-themes are thus natural experiments in the same variable: §6.1 shows
what closed-loop discovery achieves \emph{with} a trustworthy verifier;
§§6.2--6.3 show what breaks \emph{without} one.

\section{7. Foundations, Limits, and Safety of
RSI}\label{foundations-limits-and-safety-of-rsi}

The final family (60 papers) steps back from mechanisms to conditions:
when \emph{can} self-improvement continue, what bounds it, and what
follows for safety. It is the corpus's smallest family --- a
disproportion we return to in §9.

\subsection{7.1 Theoretical conditions and
bounds}\label{theoretical-conditions-and-bounds}

The clearest positive statement of conditions is Schaul's ``boundless
Socratic learning'' position paper: an agent in a closed system can
master any capability provided (a) feedback is sufficiently informative
and \emph{aligned} with the target, (b) experience coverage is broad
enough, and (c) capacity suffices --- with language games proposed as
the framework in which conditions (a) and (b) can, in principle, be
engineered \citep{schaul2024boundless}. Read against §5, the empirical
literature is one long study of what happens when (a) fails
(self-confirming loops) or (b) fails (diversity collapse); the
Socratic-learning conditions have become, in effect, the field's
implicit design checklist. On the alignment side, theoretical guarantees
for self-rewarding language models have begun to appear, characterizing
when iterative self-alignment provably improves \citep{fu2026self} ---
narrow results, but the first of their kind.

The bounding literature attacks the takeoff question from four angles.
\emph{Computability-theoretic}: a formal separation result shows that
finite internal self-modification keeps a system within its current
computational layer --- under the paper's modeling assumptions, no
amount of repeated internal revision yields the qualitative capability
jump that RSI narratives casually assume, which would require something
like stabilized access to an external oracle
\citep{lu2026computational}. The result disciplines the vocabulary:
``recursively self-improving'' and ``unboundedly self-improving'' are
different claims, and only the first is licensed by internal revision.
\emph{Dynamical}: Jafari et al.~formalize ``runaway growth'' as a
testable property, linking capability growth to resource build-out and
deriving conditions under which finite-time escalation can be ruled out
by physical and information-theoretic limits
\citep{jafari2025mathematical}. \emph{Economic}: Whitfill and Wu
estimate the elasticity of substitution between research compute and
cognitive labor from a novel panel of four frontier labs (OpenAI,
DeepMind, Anthropic, DeepSeek; 2014--2024), asking whether a
``software-only'' intelligence explosion is possible or whether compute
bottlenecks bind. Notably, their two specifications \emph{diverge}: a
baseline model estimates compute and labor as substitutes (permitting
software-only acceleration), while a ``frontier experiments'' model
accounting for the scale of state-of-the-art training runs estimates
them as complements (the bottleneck binds) --- making this the empirical
crux of the RSI-feasibility debate rather than a settled answer
\citep{whitfill2025will}. \emph{Information-theoretic}: Zenil's
impossibility result \citep{zenil2026limits} argues LLM-style
self-training \emph{cannot} be unboundedly self-improving without
symbolic model synthesis or an unvanishing stream of external signal.
Together these justify the taxonomy's central cut: bounded
self-refinement is what the theory permits without new external
resources; open-ended RSI requires either continued grounding (data,
compute, environment) or an architectural ingredient current systems
lack.

\subsection{7.2 The skeptical position}\label{the-skeptical-position}

The skeptical thread deserves separate statement because it is not a
fringe: it includes the corpus's strongest formal results. Its composite
claim: (i) self-training without external signal degrades rather than
explodes \citep{zenil2026limits, shumailov2024collapse}; (ii) even with
external signal, compute and physical constraints may prevent
super-exponential trajectories
\citep{whitfill2025will, jafari2025mathematical}; (iii) the capabilities
that would most plausibly drive takeoff --- research taste, problem
selection --- are precisely the ones current systems demonstrably lack
(§6.3). On point (iii), a first formal treatment now exists: framing
``interestingness'' --- the prospective identification of which tasks
hold potential for future progress --- as an inductive heuristic for
future compression progress, analyzable with the tools of algorithmic
information theory, and identifying it explicitly as a bottleneck on the
way to recursively self-improving systems
\citep{herrmann2026interestingness}. Research taste, in other words, is
beginning to acquire a theory, which is the precondition for it
acquiring a benchmark. Note what the skeptical position does \emph{not}
claim: none of these results bounds the impact of \emph{bounded}
self-improvement plus human direction --- Anthropic's ``compounding
efficiency'' scenario --- which is compatible with every impossibility
theorem in the corpus and is arguably just a description of current
frontier-lab practice.

\subsection{7.3 Safety, control, and the dialogue with the takeoff
question}\label{safety-control-and-the-dialogue-with-the-takeoff-question}

The safety-relevant technical literature concentrates at the category
boundaries surveyed above: permanent, self-amplifying,
population-transmissible corruption in self-evolving agent systems
\citep{lin2026safety}; capability degradation and safety drift arising
without any adversary, and the case for human-oversight anchoring
\citep{shi2026healthy}; integrity failure under completion pressure
\citep{yang2026sciintegrity}. At the architectural level, the
``unfireable safety kernel'' line argues that any control located
\emph{inside} the agent's runtime is reachable by inputs that influence
the agent --- so a system with sufficient reach into its own runtime
can, in principle, modify its own guardrails --- and that execution-time
alignment must therefore live outside the agent's address space
\citep{dobrin2026unfireable}. This is the control-theoretic restatement
of why self-modification (locus 3) changes the safety calculus:
guardrails become part of the modifiable surface.

The governance literature is thin but pointed. An analysis of inference
scaling argues the shift from pre-training compute to inference compute
may derail the current paradigm of
governance-via-training-compute-thresholds --- with opposite
implications depending on whether the inference compute is spent at
deployment or folded back into training programs inside the lab
\citep{ord2025inference}; the latter case is exactly the §4--§6 loop
structure, and it is the harder one to observe from outside.
Domain-specific evaluation frameworks for the highest-stakes
capabilities are emerging in parallel, notably for the biological
capabilities and risks of autonomous research agents, where the meaning
of an evaluation result depends on under-documented design choices
\citep{paskov2026measuring}.

Anthropic's essay \citep{anthropic2026rsi} sketches three scenarios ---
trend stall, compounding efficiency with human direction-setters, and
full RSI --- and proposes verification infrastructure for credible
slowdowns, international coordination on the arms-control model, and
urgency on alignment lest ``misalignment present in today's models
compound'' through self-improvement loops. The corpus speaks to this
framing in three ways. First, it locates the present: nearly everything
surveyed here is scenario-2 machinery --- bounded loops with
human-specified objectives --- with A-Evolve-Training
\citep{shi2026evolve} as the most scenario-3-shaped published artifact.
Second, it identifies the takeoff signal to watch: not benchmark scores
but \emph{movement up the verification hierarchy on non-verifiable
tasks} --- a system that could reliably evaluate research directions
would relax the binding constraint that currently keeps humans in the
loop, though only over the half of direction-setting that is
verification-shaped (§5.4). Third, it exposes the
misalignment-compounding concern as technically substantive rather than
speculative: the self-confirming loop (§5) is exactly the mechanism by
which a bias present in today's models would amplify under
self-training, and it is already observed at small scale
\citep{tan2025breaking}. What the technical corpus does not yet offer is
the verification infrastructure the governance proposals require ---
methods to \emph{demonstrate} that a training loop is not self-improving
past a threshold. That gap between what governance needs and what the
literature provides is, we suggest, the most underpopulated research
niche this survey has identified.

\section{8. Discussion: Cross-Cutting Observations and Open
Problems}\label{discussion-cross-cutting-observations-and-open-problems}

\textbf{The field is accelerating faster than any survey can track.}
Quarterly output in our corpus grew from single digits in early 2024 to
roughly 500 papers in 2026 Q2 (Figure 6); one theme (on-policy
self-distillation, 56 papers) did not exist as a distinct thread
eighteen months before this writing. We therefore claim durability for
this survey's \emph{structure} --- the categories-by-closure axes, the
verification hierarchy --- rather than its paper list. The structure has
already absorbed two paradigm arrivals (OPSD, zero-data self-play)
without modification, which is some evidence it will absorb the next
one. Absorption is a virtue only up to a point: a structure that
accommodates every new result predicts nothing, so §2.2 and open problem
6 state what we believe this one cannot see.

\begin{figure}
\centering
\pandocbounded{\includegraphics[keepaspectratio,alt={Figure 6: Seed-corpus quarterly growth through 2026Q2, showing log-scaled paper counts and category shares; partial 2026Q3 (data cutoff early July 2026) is omitted.}]{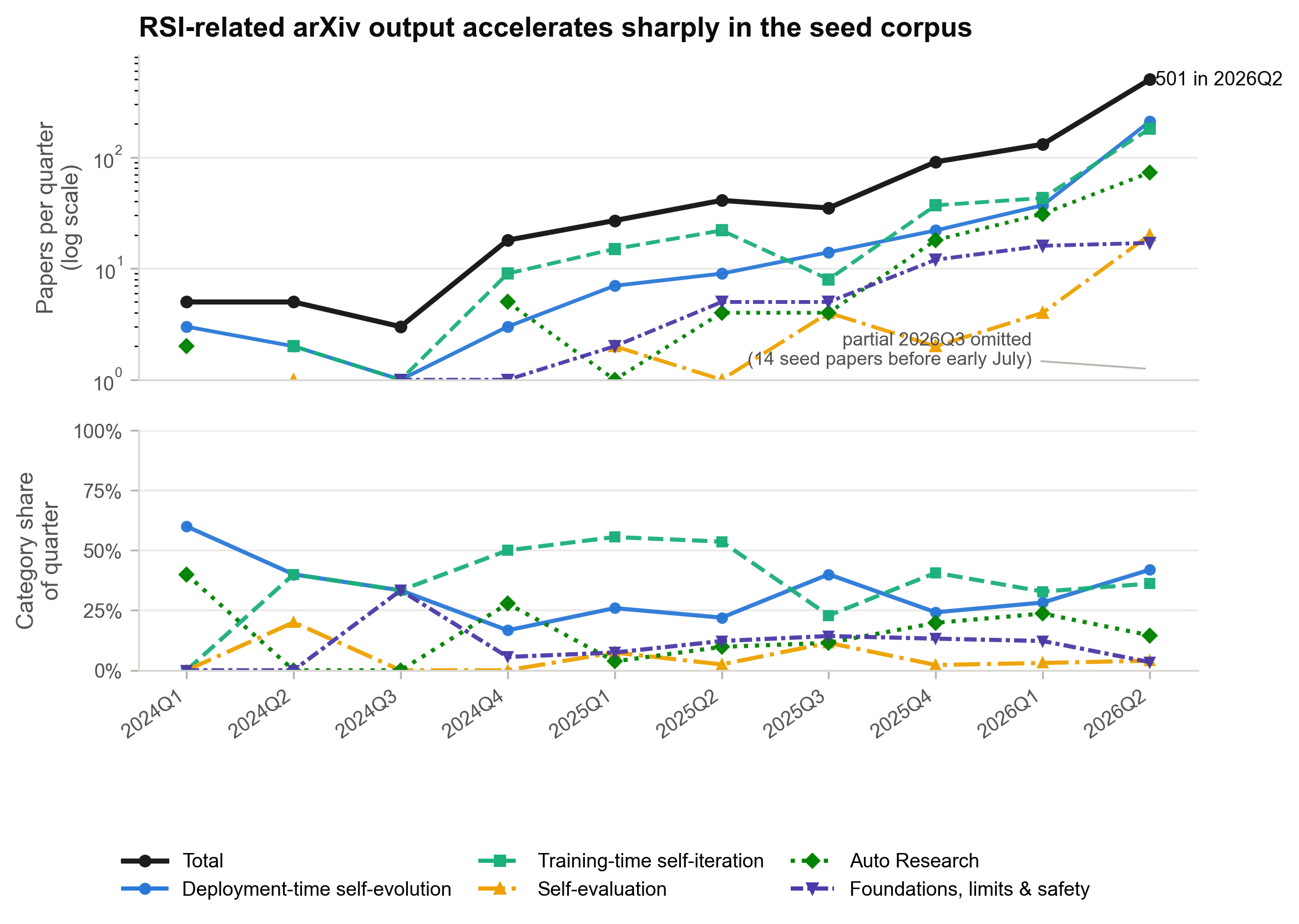}}
\caption*{Figure 6: Seed-corpus quarterly growth through 2026Q2, showing
log-scaled paper counts and category shares; partial 2026Q3 (data cutoff
early July 2026) is omitted.}
\end{figure}

\textbf{The corpus's shape itself asks for explanation.} The category
proportions in Table 1 and the share panel of Figure 3 are conspicuously
uneven, and it is worth being explicit about both what that unevenness
can and cannot mean. It cannot be read directly as the field's true
composition: our seed queries and deliberately targeted supplement shape
the proportions (§2.3), so the raw counts partly mirror our own
sampling. What survives that caveat, because it is visible \emph{within}
comparable samples, still calls for explanation: deployment- and
training-time work dominates every seed quarter; Auto Research is a
factor of two to three smaller despite carrying the field's most
celebrated results; foundations is smallest by far. We offer four
non-exclusive hypotheses, testable in principle. \emph{A verifiability
gradient}: the two dominant categories concentrate where cheap verifiers
exist (code, math), so experiments are fast and publishable --- the same
gradient §5.2 documents inside methods reappears as a force on topic
selection. \emph{A cost-of-entry gradient}: an inference-time refinement
study needs API credits; a training-loop paper needs GPUs; an
Auto-Research system needs infrastructure, long horizons, and (for
frontier results) lab-scale compute --- paper counts fall roughly as
entry costs rise, and the compute concentration documented for frontier
labs \citep{whitfill2025will} pushes the far end of the spectrum behind
corporate walls, where §2.3's publication-censoring effect hides it from
any arXiv sample. \emph{Herd dynamics}: the OPSD thread crystallizing
from nothing to dozens of papers within months (§4.3) shows how fast
attention concentrates once a recipe is legible and benchmarkable;
legibility, not importance, sets the gradient. \emph{Institutional role
assignment}: evaluation was until recently a service function inside
method papers rather than a research identity --- our supplement had to
target it precisely because the self-improvement literature treats it as
plumbing --- and the foundations deficit (60 of 1,250, against the
stakes claimed in §7) plausibly reflects that theory and governance work
is rewarded by neither benchmarks nor product roadmaps. We flag these as
hypotheses rather than findings; distinguishing them would itself be a
useful study, and the data to start it ships with this paper.

\textbf{Mechanisms outrun evaluation.} The ``self-X'' vocabulary keeps
growing, but the papers our analysis leaned on hardest were
disproportionately diagnostic: controlled decompositions of feedback
value \citep{cupia2026drives}, placebo-controlled self-repair studies
\citep{iscan2026falsification}, stability analyses of self-play
\citep{pu2026survive}, integrity benchmarks
\citep{yang2026sciintegrity}. This critique literature is no longer a
reaction to the field; it is a load-bearing part of it, and in several
themes (AI scientists most clearly) it is the better-cited part. We read
this as a maturation signal --- the same sequence (method wave, then
measurement wave) that reasoning research went through in 2023--2024.

\textbf{The field's emerging answer to the verification bottleneck is
evaluator co-evolution.} Across otherwise unconnected themes, 2026
produced the same architectural move independently at least five times:
co-evolving self-generated unit tests with the code they judge
\citep{hu2026cosplay}, discovering per-judge competence taxonomies
before trusting judge scores \citep{cao2026each}, decomposing opaque
judges into auditable binary questions \citep{cho2026ask}, self-training
the verifier as the primary object of improvement \citep{wu2026self},
and making the evaluation criterion itself part of the evolutionary loop
\citep{iacob2026red, zhang2026scientific}. The field has evidently
concluded that static verifiers cannot supervise improving systems ---
the Red Queen framing --- and is betting that the verifier must improve
alongside the policy. Whether this escapes the self-confirming loop or
merely gives it a second story is, in our view, the pivotal empirical
question the next two years will answer; the A-Evolve-Training episode
(§6.2), where a loop detected its own proxy's corruption and revised its
search policy, is the first field observation suggesting escape is
possible.

\textbf{Modality generalization is real but signal-limited.} The
paradigm has spread from text to vision-language, video, robotics, and
speech (§§3.3, 4.5). But each port re-encounters the verification
hierarchy at a lower rung: execution feedback has no cheap embodied
analogue, and multimodal self-consistency signals admit shortcuts that
text-only loops do not \citep{wang2026seeing, venkatraman2026paying}.
The generalization frontier is therefore not model capability but
\emph{signal engineering} --- physics residuals
\citep{hassan2026proprio}, formal skill verification
\citep{yang2026vaso} --- modality by modality.

\textbf{Open problems.} Six, in rough order of leverage:

\begin{enumerate}
\def\labelenumi{\arabic{enumi}.}
\tightlist
\item
  \textbf{The exchange rate of grounding.} Theory says pure closed loops
  degrade \citep{zenil2026limits}; practice mixes in external signal ad
  hoc. Nobody has characterized the minimum exogenous-signal rate that
  sustains improvement, though the entropy-reservoir framework
  \citep{jingwei2025entropy} suggests the question is well-posed.
\item
  \textbf{Verifying the non-verifiable.} Progress on research taste,
  creative quality, and direction-setting evaluation would relax the
  binding constraint of the entire spectrum (§5). Meta-evaluation
  \citep{sui2026conversation} and auditability-by-construction
  \citep{rasheed2026fluent} are starts, not solutions.
\item
  \textbf{Stability engineering as a discipline.} Rise-and-collapse
  \citep{lin2026self}, diversity collapse \citep{dineen2026vocabulary},
  and safety drift \citep{shi2026healthy} are currently rediscovered
  per-paradigm. A unified treatment of self-improvement loop dynamics
  --- when they converge, oscillate, or collapse --- would replace
  today's per-theme folklore.
\item
  \textbf{Trustworthy accumulation.} Persistent self-modification
  (skills, memory, experience graphs) lacks the verification story that
  weights-based training has. The formal-verification approach
  \citep{yang2026vaso} and the threat analyses \citep{lin2026safety}
  bracket the problem from two sides; the middle is open.
\item
  \textbf{Governance-grade measurement.} The verification infrastructure
  that credible slowdown proposals require \citep{anthropic2026rsi} ---
  auditable evidence about what a training loop is and is not improving
  --- has essentially no technical literature behind it. Given that the
  foundations family (§7) is the corpus's smallest at 60 of 1,250
  papers, the mismatch between the stated stakes and the allocated
  research effort is the clearest gap this survey has found.
\item
  \textbf{Recognizing frame revision.} Every problem above is posed
  inside a fixed objective. The prior question --- whether a system can
  notice that the objective itself has gone stale, and reframe while
  preserving what should survive the reframing --- has neither a
  benchmark nor a measurement convention, and is invisible to the
  instrument this survey builds (§2.2). It is not obviously
  unmeasurable. Four observables are candidates: \emph{productive
  abandonment} (dropping a line whose local metric is still improving,
  scored against later vindication); \emph{invariant transfer} (which
  constraints a system carries across a reformulation of its own task);
  \emph{absence-sensitivity} (whether a missing result is treated as
  evidence --- testable against the null-result deficit formalized in
  \citep{chauhan2026dead}); and \emph{reframing latency} (how much
  evidence a loop needs to abandon a stale objective after its
  environment shifts). A benchmark of deliberately stale or misspecified
  objectives, scored on whether a system notices rather than on how well
  it optimizes, is buildable today and does not exist. Part of the
  obstacle is that the record deletes the evidence: publication discards
  failed experiments, rejected hypotheses, and the branching exploration
  itself --- the ``storytelling tax'' that agent-native artifact
  protocols are designed to reverse \citep{liu2026last}. We note the
  reflexive point: this section argues that legibility, not importance,
  sets the field's attention gradient; a taxonomy is a legibility
  instrument, and ours is subject to the same critique.
\end{enumerate}

\section{9. Conclusion}\label{conclusion}

We surveyed 1,250 recent papers on AI self-improvement through a
two-axis taxonomy --- what the system improves (outputs, policy,
scaffolding, the research process) and who validates the improvement ---
and argued that the axis structure resolves what the ``self-X''
vocabulary obscures: bounded self-refinement and open-ended recursive
self-improvement are different phenomena with different evidence bases,
different theory, and different risk profiles.

The evidence sorts cleanly. Bounded self-refinement is an engineering
success: inference-time loops reliably improve outputs when grounded in
external signals; training-time loops persist those gains and are
industrial practice; agents accumulate skills and experience across
episodes; and evolutionary discovery systems produce artifacts ---
algorithms, kernels, mathematical constructions --- that feed back into
AI development itself. Open-ended RSI, by contrast, remains bounded on
every side we can measure: theoretically by grounding requirements and
compute elasticities, empirically by collapse dynamics, and practically
by the non-verifiability of exactly the judgments (what to work on, what
counts as better) that would make the loop self-sufficient.

These are not one difficulty. ``What counts as better'' is a
verification problem the hierarchy indexes; ``what to work on'' is prior
to it, and this survey's instrument --- like the field's --- measures
the first far better than the second.

The single thread connecting every category is that self-improvement is
only as real as its verification. That framing converts the takeoff
question from speculation into a measurement program: watch not
benchmark scores but the verification hierarchy --- in particular,
whether systems become reliable evaluators of open-ended research
judgment. Until then, the human role in the loop is not a sentimental
holdover; it is the field's verification layer of last resort. The
literature surveyed here suggests that layer will be load-bearing for
some time --- and that building its eventual replacement, carefully and
measurably, is the research program this decade's AI safety may most
depend on.

\section{Data availability}\label{data-availability}

The full 1,250-paper corpus with taxonomy category assignments (plus the
unsupervised 13-topic clustering that organized the seed harvest before
the taxonomy of §2.2 was fixed), the per-category statistics underlying
Table 1, the figure sources, and the scripts that generate the
bibliography and figures are available at
\url{https://github.com/deepgrounding/recursive-self-improvement}.

\section{References}\label{references}

\bibliography{references}

\end{document}